\documentclass[preprint]{imsart}

\usepackage{graphicx} 
\usepackage{xcolor}

\usepackage{amsfonts,amssymb,float}

\DeclareSymbolFont{sfletters}{OML}{cmbrm}{m}{it}

\DeclareMathSymbol{\salpha}{\mathord}{sfletters}{"0B}

\usepackage{mathrsfs}
\RequirePackage{natbib}
\RequirePackage[colorlinks,citecolor=blue,urlcolor=blue]{hyperref}

\usepackage{xr}
\usepackage{xcite}
\usepackage{overpic}
\makeatletter
\newcommand{\oset}[3][1ex]{%
  \mathrel{\mathop{#3}\limits^{
    \vbox to#1{\kern-2\ex@
    \hbox{$\scriptstyle#2$}\vss}}}}
\makeatother

\usepackage{mathtools}
\RequirePackage{natbib}
\usepackage{comment}
\usepackage{enumerate}
\usepackage{eufrak}

\usepackage{array,varwidth}

\usepackage{overpic}
\usepackage[framemethod=tikz,innertopmargin=7pt]{mdframed}
\usepackage{graphicx} 
\usepackage{subfig}
\usepackage{placeins} 

\RequirePackage[OT1]{fontenc}
\RequirePackage{amsthm,amsmath}

\startlocaldefs
\numberwithin{equation}{section}
\theoremstyle{plain}
\newtheorem{theorem}{Theorem}[section]
\newtheorem{lemma}{Lemma}[section]
\newtheorem{proposition}{Proposition}[section]

\usepackage{algorithm}
\usepackage{algorithmic}

\theoremstyle{remark}

\newcommand{\R}{\mathbb{R}}

\renewcommand{\hat}[1]{\widehat{#1}}

\renewcommand{\L}{\mathcal{L}}
\newcommand{\bxi}{\boldsymbol \xi}

\newcommand{\Mse}{\textsc{mse}}
\newcommand{\Err}{\textsc{err}}

\newcommand{\mse}{\text{mse}}

\newcommand{\tq}{{\tt{q}}}

\newcommand{\VI}{\textsc{vi}}

\newcommand{\oob}{\textsc{oob}}

\newcommand{\D}{\mathcal{D}}

\newcommand{\X}{\mathcal{X}}

\renewcommand{\P}{\mathbb{P}}
\newcommand{\E}{\mathbb{E}}

\renewcommand{\L}{\mathcal{L}}

\newcommand{\e}{\epsilon}
\newcommand{\ve}{\varepsilon}

\newcommand{\var}{\operatorname{var}}

\newcommand{\tsum}{\textstyle\sum}
\newcommand{\ts}{\textstyle}

\endlocaldefs

\usepackage[T1]{fontenc}
\begin{document}

\begin{frontmatter}
\title{Measuring the Algorithmic Convergence of Randomized Ensembles:\\The Regression Setting}

\begin{aug}
~\\
\author{\fnms{Miles} \snm{E.  Lopes}\thanksref{t1}\ead[label=e1]{???}} \ \ \
\author{\fnms{Suofei} \snm{Wu}} \ \ \
\author{\fnms{Thomas} \snm{C. M. Lee}\thanksref{t3}
\ead[label=e3]{???}}

\thankstext{t1}{Supported in part by NSF grant DMS 1613218.}
\thankstext{t3}{Supported in part by NSF grants DMS 1811405 and DMS 1811661.}

\affiliation{University of California, Davis}

\end{aug}

\begin{aug}
\address{\normalsize University of California, Davis}
%
\end{aug}

\begin{abstract}
When randomized ensemble methods such as bagging and random forests are implemented, a basic question arises: Is the ensemble large enough? In particular, the practitioner desires a rigorous guarantee that a given ensemble will perform nearly as well as an ideal infinite ensemble (trained on the same data).  The purpose of the current paper is to develop a bootstrap method for solving this problem in the context of regression --- which complements our companion paper in the context of classification~\citep{Lopes:2019}. In contrast to the classification setting, the current paper shows that theoretical guarantees for the proposed bootstrap can be established under much weaker assumptions. In addition, we illustrate the flexibility of the method by showing how it can be adapted to measure algorithmic convergence for variable selection. Lastly, we provide numerical results demonstrating that the method works well in a range of situations. \end{abstract}

\begin{keyword}[class=MSC]
\kwd[Primary]  { 62F40 }
\kwd[secondary] { 65B05, 68W20, 60G25 } 
\end{keyword}
\begin{keyword}
\kwd{random forests, bagging, bootstrap, randomized algorithms}
\end{keyword}

\end{frontmatter}

\section{Introduction}
Ensemble methods are a fundamental approach to prediction, based on the principle that accuracy can be enhanced by aggregating a diverse collection of prediction functions.
Two of the most widely used methods in this class are  \emph{random forests} and \emph{bagging}, which rely on randomization as a general way to diversify an ensemble~\citep{breiman1996,breiman2001}. For these types of randomized ensembles, it is generally understood that the predictive accuracy improves and eventually stabilizes as the ensemble size becomes large. Likewise, in the theoretical analysis of randomized ensembles, it is common to focus on the idealized case of an infinite ensemble~\citep{Buhlmann:Yu,hallsamworth,biau,biau2012,scornetconsistency}. However, in practice, the user does not know the true relationship between accuracy and ensemble size, and as a result, it is difficult to know if an ensemble is sufficiently large. 

The purpose of the current paper is develop a solution to this problem for random forests, bagging, and related methods in the context of regression. More specifically, we offer a bootstrap method for estimating how far the prediction error of a finite ensemble is from the ideal prediction error of an infinite ensemble (trained on the same data). 
A precise description of the setup and problem formulation is given as follows.

\subsection{Background and setup}\label{sec:setup}
To fix some basic notation for the regression setting, let $\D=\{(X_j,Y_j)\}_{j=1}^n$ denote a set of training data in  a space $\mathcal{X}\times \R$, where each $Y_j$ is the scalar response variable associated to $X_j$, and the space $\mathcal{X}$ is arbitrary. Also, an ensemble of $t$ regression functions trained on $\D$ is denoted as $T_i: \mathcal{X}\to\R$, where $i=1,\dots,t$, and the number $t$ is referred to as the ensemble size.
\paragraph{Randomized regression ensembles.} For the purpose of understanding  our setup, it is helpful to quickly review the methods of bagging and random forests. The method of bagging works by generating random sets $\D_1^*,\dots,\D_t^*$, each of size $n$, by sampling with replacement from $\D$. Next, a standard ``base'' regression algorithm is used to train a regression function $T_i$ on $\D_i^*$ for each $i=1,\dots,t$. For instance, it is especially common to apply a decision tree algorithm like CART~\citep{CART} to each set $\D_i^*$. In turn,  future predictions are made by using the averaged regression function, which is defined for each $x\in\mathcal{X}$ by
\begin{equation}\label{eqn:Tbardef}
 \bar T_t(x):=\frac 1 t\sum_{i=1}^t T_i(x).
 \end{equation}
Much like bagging, the method of random forests uses sampling with replacement to generate the same type of random sets $\D_1^*,\dots,\D_t^*$. However, random forests adds an additional source of randomness when the base regression algorithm is applied to each $\D_i^*$. Namely, in the standard case when $\mathcal{X}\subset\R^p$ and CART is the base regression algorithm, random forests 
uses randomly chosen subsets of the $p$ features when ``split points'' are selected for the CART regression trees. Likewise, random forests also uses the average~\eqref{eqn:Tbardef} when making final predictions. A more detailed description may be found in~\cite{ESL}.

In order to unify the methods of bagging and random forests within a common theoretical framework, our analysis will consider a more general class of randomized ensembles. This class consists of regression functions $T_1,\dots,T_t$ that can be represented in the abstract form
\begin{equation}\label{eqn:rep}
T_i(x) = \varphi(x;\D,\xi_i),
\end{equation}
where $\xi_1,\dots,\xi_t$ are i.i.d.~``randomizing parameters'' generated independently of $\D$, and $\varphi$ is a deterministic function that does not depend on $n$ or $t$. In particular, the representation~\eqref{eqn:rep} implies that the random functions $T_1,\dots,T_t$ are conditionally i.i.d., given $\D$. To see why bagging is representable in this form, note that $\xi_i$ can be viewed as a random vector that specifies which points in $\D$ are randomly sampled into $\D_i^*$. Similarly, in the case of random forests, each $\xi_i$ encodes the points in $\D_i^*$, as well as randomly chosen sets of features used for training $T_i$. More generally, the representation~\eqref{eqn:rep} is relevant to other types of randomized ensembles, such as those based on random rotations~\citep{Blaser:2016}, random projections~\citep{cannings2017}, or posterior sampling~\citep{NgJordan,chipman2010bart}.

\paragraph{Algorithmic convergence.}  In our analysis of algorithmic convergence, we will focus on quantifying how the mean-squared error (MSE) of an ensemble behaves as the ensemble size $t$ becomes large. To define this measure of error in more precise terms, let $\bxi_t:=(\xi_1,\dots,\xi_t)$ denote the randomizing parameters of the ensemble, and let $\nu=\mathcal{L}(X,Y)$ denote the joint distribution of a test point $(X,Y)\in\X\times \R$, which is drawn independently of $\D$ and $\bxi_t$. Accordingly, we define
\begin{equation}\label{eqn:msedef}
\Mse_t \, := \, \int_{\X\times \R} \big( y-\bar T_t(x)\big)^2d\nu(x,y) \, = \, \E\Big[(Y-\bar T_t(X))^2\,\Big|\,\boldsymbol \xi_t, \D\Big],
\end{equation}
where the expectation on the right is only over the test point $(X,Y)$.
In this definition, it is important to notice that $\Mse_t$ is a random variable that depends on  both $\bxi_t$ and $\D$. However, due to the fact that the \emph{algorithmic} fluctuations of $\Mse_t$ arise only from $\bxi_t$, we will view the set $\D$ as a fixed input to the training algorithm, and likewise, our analysis will always be conditional on $\D$. Indeed, the conditioning on $\D$ is motivated by the fact that the user would like to assess convergence for the particular set $\D$ that they actually have, and this approach has been adopted in several other analyses of algorithmic convergence for randomized ensembles~\citep{NgJordan,lopes2016,scornet2016asymptotics,cannings2017, Lopes:2019}.

\begin{figure}[H]
	\begin{minipage}{\textwidth}
		\begin{centering}
			
			\subfloat{
				\begin{overpic}[width=50mm,height=50mm,angle=270]{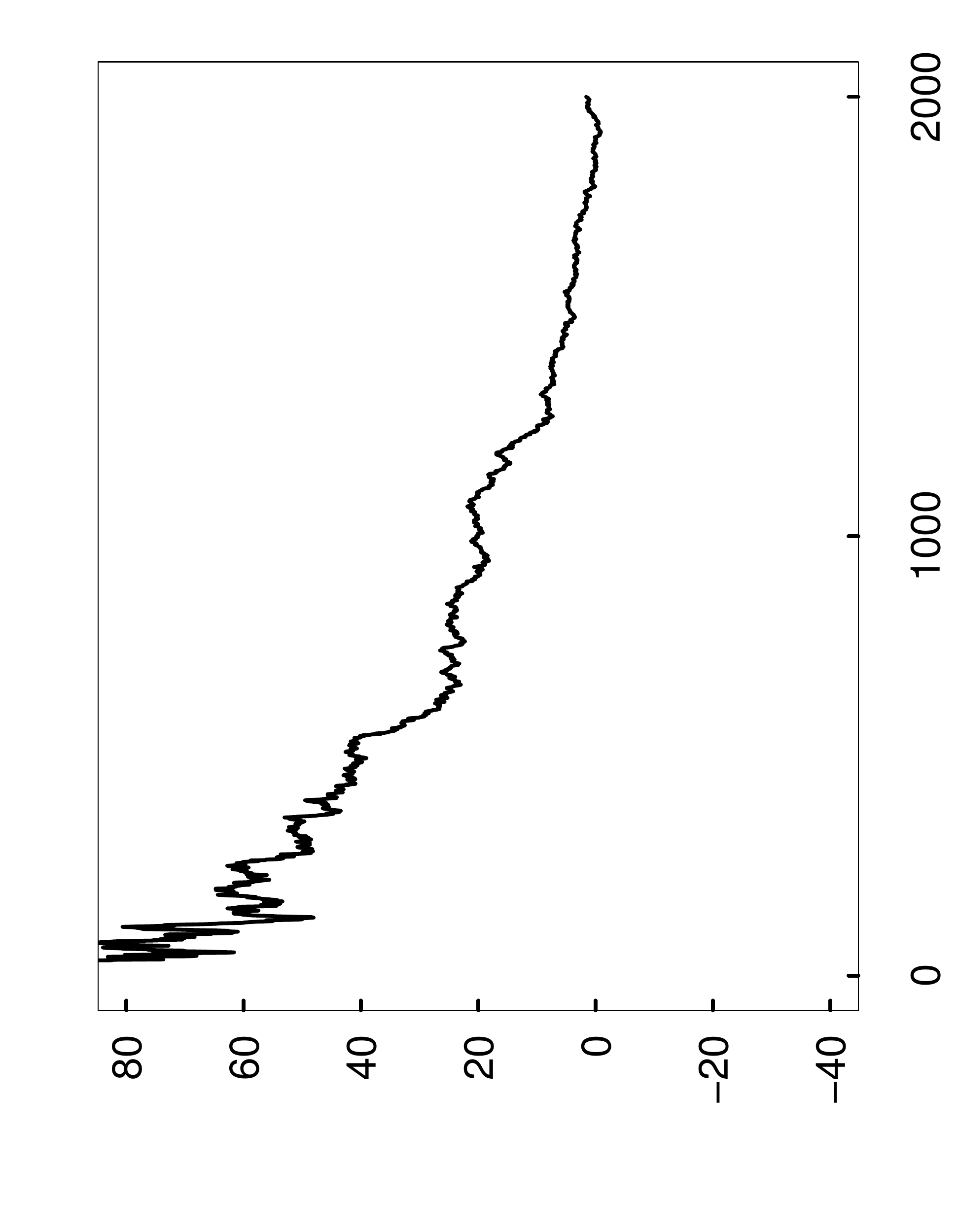} 
					\put(40,-7){\color{black}{\small  ensemble size $t$}}   
					\put(-3,35){\rotatebox{90}{\small  $\Mse_t-\mse_{\infty}$}}
				\end{overpic}	
			}  \ \ \ \ \ \ \ \
			\subfloat{
				\begin{overpic}[width=50mm,height=50mm,angle=270]{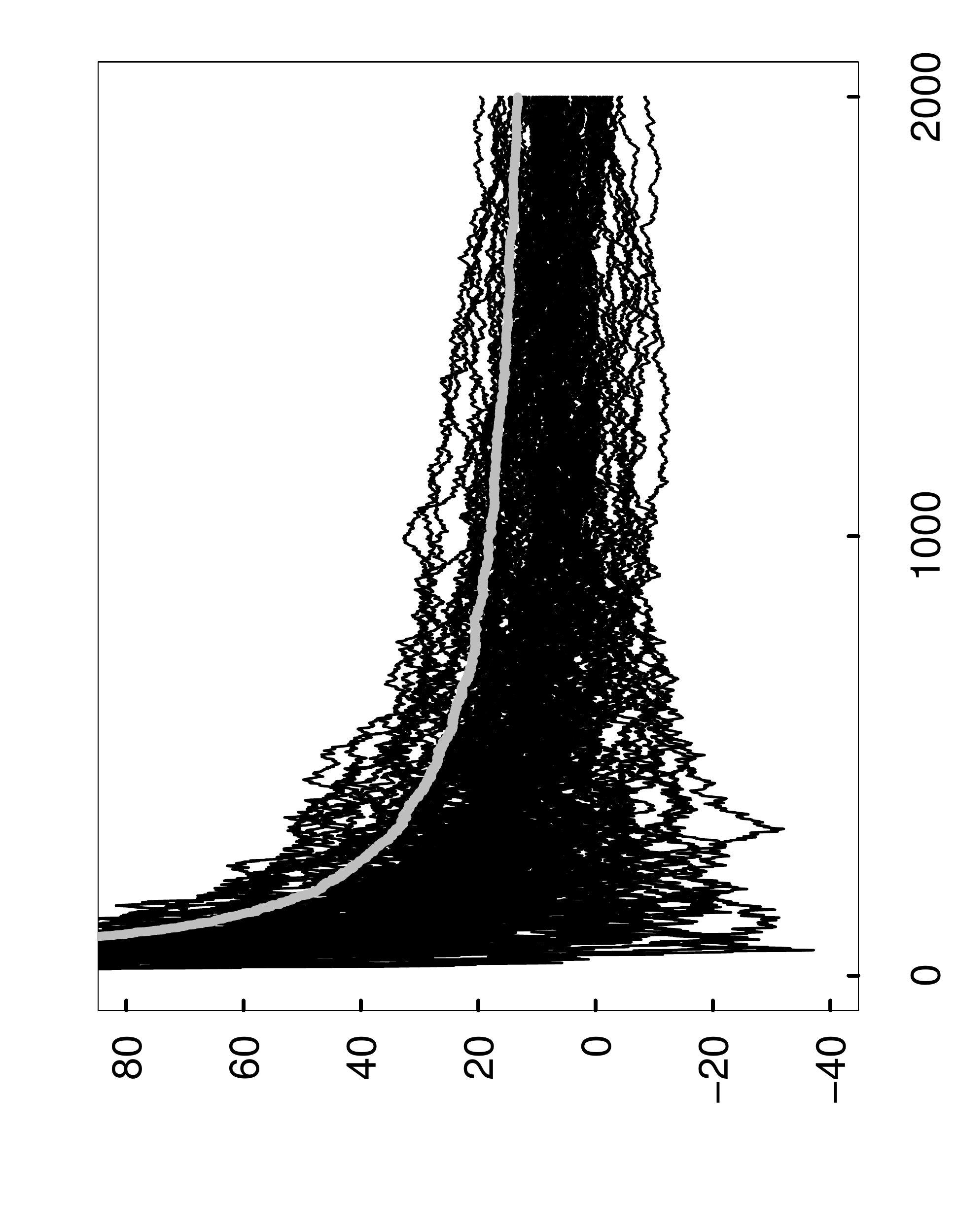} 
					\put(40,-7){\color{black}{\small ensemble size $t$}}   
					\put(-3,35){\rotatebox{90}{\small $\Mse_t-\mse_{\infty}$}}
					\put(55,65){\color{gray}{\Huge -}}
					\put(63,68){\color{black}{\small $q_{0.9}(t)$}}
				\end{overpic}	
			} 
			\par\end{centering}
		\vspace{0.3cm}
		\caption{Left Panel: A sample path of $\Mse_t-\mse_{\infty}$ over a single run of random forests on the Housing  Data described in Section~\ref{sec:expt}. Right Panel: 1,000 sample paths of $\Mse_t-\mse_{\infty}$, with $q_{.90}(t)$ overlaid in gray. In addition, note the curves in the two panels are not observable in practice, and they are presented only for illustration.
			\vspace{-0.0cm}
		}

		\label{fig:raw-data}
	\end{minipage}
\end{figure}
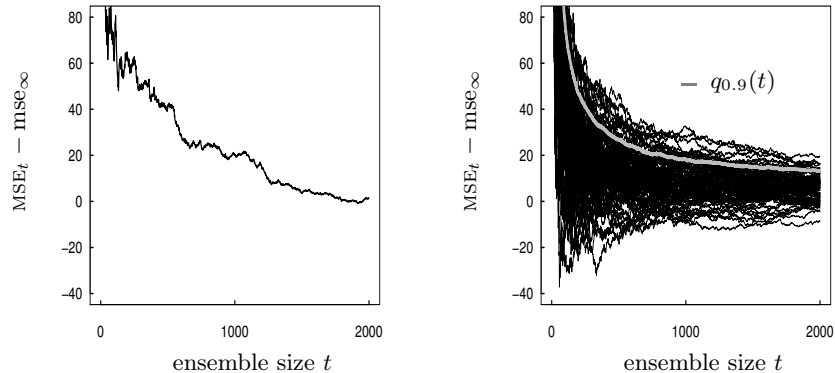

As a way of illustrating algorithmic convergence, Figure~\ref{fig:raw-data} shows how $\Mse_t$ evolves when the random forests method is applied to a fixed training set $\D$. More specifically, if $\mse_{\infty}$ denotes the limit of $\Mse_t$ as $t\to\infty$, then the left panel displays successive values of the convergence gap $\Mse_t-\mse_{\infty}$ as decision trees are added during a single run of random forests, from $t=1$ up to $t=2,\!000$. After this entire process is repeated 1,000 times on the same set $\D$, we obtain many overlapping sample paths, as shown in the right panel of Figure~\ref{fig:raw-data}. (Note also that none of these curves are observable in practice, and the figure is given only for illustration.)

From a practical standpoint, the user would like to know the size of the convergence gap $\Mse_t-\mse_{\infty}$ as a function of $t$. For this purpose, it is useful to consider the $(1-\alpha)$-quantile of $\Mse_t-\mse_{\infty}$, which is defined for any $\alpha\in(0,1)$ by
\begin{equation*}
q_{1-\alpha}(t) \, := \, \inf\Big\{q\in\R\,\Big|\, \P\big(\Mse_t-\mse_{\infty}\leq q\,\big|\,\D\big) \, \geq \, 1-\alpha \Big\}.
\end{equation*}
In other words, the value $q_{1-\alpha}(t)$ is the \emph{tightest possible} upper bound on the gap  that holds with probability at least $1-\alpha$, conditionally on the set $\D$. This interpretation of $q_{1-\alpha}(t)$ can also be understood from the right panel of Figure~\ref{fig:raw-data}, where we have plotted $q_{1-\alpha}(t)$ in gray, with $\alpha=1/10$.

\paragraph{The problem to be solved.} Although it is clear that the quantile $q_{1-\alpha}(t)$ represents a precise measure of algorithmic convergence, this function is unknown in practice. This leads to the problem of estimating $q_{1-\alpha}(t)$, which we propose to solve.

Beyond the fact that $q_{1-\alpha}(t)$ is unknown, it is also important to keep in mind that estimating $q_{1-\alpha}(t)$ involves some additional constraints. First, the user would like to be able to assess convergence from the output a \emph{single} run of the ensemble method, whereas the function $q_{1-\alpha}(t)$ describes the fluctuations of $\Mse_t-\mse_{\infty}$ over \emph{repeated} runs, as illustrated in the right panel of Figure~\ref{fig:raw-data}. Hence, at first sight, it is not obvious that the output of a single run provides enough information to successfully estimate $q_{1-\alpha}(t)$.
 Second, the method for estimating $q_{1-\alpha}(t)$ should be computationally inexpensive, so that the cost of checking convergence is manageable in comparison to the cost of training the ensemble itself. Later on, we will show that the proposed method is able to handle both of these constraints, in Sections~\ref{sec:method} and~\ref{sec:comp} respectively.

		\vspace{0.1cm}

	\subsection{Related work and contributions} 
	\vspace{0.1cm}
	The general problem of measuring the algorithmic convergence of randomized ensembles has attracted sustained interest over the past two decades. In particular, there have been numerous empirical studies of algorithmic convergence for both classification and regression~\citep[e.g.][]{latinne,comet,swiss,oshiro,probst2018}. 
	
	With regard to the theoretical analysis of convergence, we will now review the existing results for classification and regression separately. In the setting of classification, much of the literature has studied convergence in terms of the misclassification probability for majority voting, denoted $\Err_t$ (a counterpart of $\Mse_t)$, which is viewed as a random variable that depends on $\bxi_t$ and $\D$. For this measure of error, the convergence of $\E[\Err_t|\D]$ and $\var(\Err_t|\D)$ as $t\to\infty$ has been analyzed in the papers~\citep{NgJordan,lopes2016,cannings2017}, which have developed asymptotic formulas for $\E[\Err_t|\D]$, as well as bounds for $\var(\Err_t|\D)$. Related results for a different measure of error can also be found in~\cite{hernandez}. More recently, our companion paper~\citep{Lopes:2019} has developed a bootstrap method for measuring the convergence of $\Err_t$, which is able to circumvent some of the limitations of analytical results.

	In the setting of regression, algorithmic convergence results on $\Mse_t$ are scarce in comparison to those for $\Err_t$. Instead, much more attention in the regression literature has focused on how the size of $t$ influences the variance of point predictions $\bar T_t(x)$, with $x\in\X$ held fixed~\citep[e.g.,][]{sexton2009standard,arlot2014,wager2014confidence,mentch2016,scornet2016asymptotics}.  To the best of our knowledge, the only paper that has systematically studied algorithmic convergence in terms of an error measure is~\citep{scornet2016asymptotics}, which considers the risk $r_t:=\E[(\bar T_t(X)-\mu(X))^2]$, where $\mu(x):=\E[Y|X=x]$ is the true regression function, and the expectation in the definition of $r_t$ is over  $(X,\D,\bxi_t)$. In particular, the paper~\citep{scornet2016asymptotics} develops an elegant non-asymptotic bound on the gap between $r_t$ and its limiting value $r_{\infty}$ as $t\to\infty$. Under the assumption of  a Gaussian regression model with $\mathcal{X}=[0,1]^p$, this bound has the form

	\vspace{0.1cm}
\begin{equation}\label{eqn:scornetbound}
 r_t-r_{\infty} \, \leq \, \ts\frac{8}{t}\Big(\|\mu\|_{\infty}^2+\sigma^2(1+4\log(n))\Big),
 	\vspace{0.1cm}
 \end{equation}
where $\sigma^2=\var(Y)$, and $\|\mu\|_{\infty}:=\sup_{x\in\mathcal{X}}|\mu(x)|$. In addition to this bound, the paper~\citep{scornet2016asymptotics} gives further insight into algorithmic convergence by developing a precise uniform central limit theorem for $\bar T_t$ as $t\to\infty$, with $\D$ held fixed. More specifically, this limit theorem demonstrates that under certain conditions, the standardized process $\sqrt{t}(\bar T_t(\cdot)-\E[\bar T_t(\cdot)|\D])$ converges in distribution (conditionally on $\D$) to a Gaussian process on $\mathcal{X}$. 

\vspace{0.1cm}
\paragraph{Contributions.} From a methodological standpoint, the approach taken here differs in several ways from previous works in the regression setting. Most notably, our work looks at algorithmic convergence in terms of an error measure that is conditional on $\D$. (For instance, this differs from the analysis of $r_t$, which averages over $\D$.) In particular, we provide a quantile estimate $\hat{q}_{1-\alpha}(t)$, such that the bound
\vspace{-0.1cm}
\begin{equation*}
\Mse_t-\mse_{\infty} \, \leq \  \hat q_{1-\alpha}(t)
\end{equation*}
holds with a probability that is effectively $1-\alpha$, conditionally on $\D$. This conditioning is especially important from the viewpoint of the user, who is typically interested in convergence with respect to the \emph{actual dataset at hand}. 
Another distinct feature of our method is that it provides the user with a \emph{direct numerical estimate of convergence}, whereas formula-based results  are more likely to involve conservative constants, or depend on unknown parameters, such as $\|\mu\|_{\infty}$ or $\sigma$ in the bound~\eqref{eqn:scornetbound}.

In addition, the scope of the proposed method goes beyond $\Mse_t$, and in Section~\ref{sec:VI} we will show how the bootstrap method is flexible enough that it can also be applied to variable selection. In this context, the ensemble provides a ranking of variables according to an ``importance measure'', and this ranking typically stabilizes as $t\to\infty$. However, the notion of convergence is somewhat subtle, because it is possible that the importance measure for some variables may converge more slowly than for others --- \emph{which can distort the overall ranking of variables}. As far as we know, this issue has not be addressed in the literature, and the method proposed in Section~\ref{sec:VI} provides a way to check that convergence has been achieved uniformly across all variables, so that they can be compared fairly.

With regard to theory, the most important aspects of our analysis is that it is based on very mild assumptions.  To place our assumptions into context, it is worth emphasizing that most analyses of randomized ensembles deal with specialized types of prediction functions $T_1,\dots,T_t$ that are much simpler than the ones used in practice~\citep[e.g.][]{linjeon,arlot2014,biau,biau2012,scornetconsistency,scornet2016asymptotics,scornetIEEE,Lopes:2019}. By contrast, our current results for regression only rely on the representation~\eqref{eqn:rep} and basic moment assumptions (to be detailed in Section~\ref{sec:main}). In particular, the crucial ingredient that enables us to handle general types of prediction functions is a version of Rosenthal's inequality due to~\cite{Talagrand:1989}, which is applicable to sums of independent Banach-valued random variables. Moreover, this  allows our analysis to be fully \emph{non-asymptotic}.

\paragraph{Outline.}	The remainder of the paper is organized as follows. The proposed methods are described in Section~\ref{sec:method}, and our main result on bootstrap consistency is presented in Section~\ref{sec:main}. Next, the computational cost of the methods is assessed in Section~\ref{sec:comp}, and numerical experiments are given in Section~\ref{sec:expt}. Finally, all proofs are given in the supplementary material.

	\section{Methodology}\label{sec:method}
	
	Below, we present our core method for measuring algorithmic convergence with respect to $\Mse_t$ in Section~\ref{sec:alg1}. Next, we show how this approach can be extended to measuring convergence with respect to variable importance in Section~\ref{sec:VI}.
	
	\subsection{Measuring convergence with respect to mean-squared error}\label{sec:alg1}
	The intuition for the proposed method is based on two main considerations. First, the definition of $\Mse_t$ in equation~\eqref{eqn:msedef} shows that it can be interpreted as a functional of $\bar T_t$. More specifically, if we let $f:\mathcal{X}\to\R$ denote a generic function, then we define the functional $\psi$ according to
	\begin{equation}
	\psi(f) =\int_{\mathcal{X}\times \R}(y-f(x))^2d\nu(x,y),
	\end{equation}
and it follows that $\Mse_t$ can be written as
	\begin{equation}\label{eqn:mserep}
	\Mse_t \, = \, \psi(\bar T_t).
	\end{equation}
Second, it is a general principle that bootstrap methods are well-suited to approximating distributions derived from smooth functionals of sample averages --- which is precisely what the representation~\eqref{eqn:mserep} entails.

To make a more direct connection between these general ideas and the problem of estimating $q_{1-\alpha}(t)$, recall that we actually need to approximate the distribution of the difference $\Mse_t-\mse_{\infty}$, rather than just $\Mse_t$ itself. Fortunately, the limiting value $\mse_{\infty}$ can be linked with $\psi$ through function 
\begin{equation}\label{eqn:varthetadef0}
\vartheta(x):=\E[\bar T_t(x)|\D],
\end{equation}
where the expectation is only over the algorithmic randomness in $\bar T_t$ (i.e.~over the random vector $\bxi_t$).
More specifically, when the functions $T_1,\dots,T_t$ satisfy the representation~\eqref{eqn:rep}, the law of large numbers implies
$\mse_{\infty}=\psi(\vartheta)$ under basic integrability assumptions,
which leads to the relation 
\begin{equation}
\Mse_t-\mse_{\infty} \, = \, \psi(\bar T_t)-\psi(\vartheta).
\end{equation}
This relation is the technical foundation for the proposed method, since it suggests that in order to mimic the fluctuations of $\Mse_t-\mse_{\infty}$, we can develop a bootstrap method by viewing the functions $T_1,\dots, T_t$ as ``observations'', and viewing $\bar T_t$ as an estimator of $\vartheta$. In other words, if we sample $t$ functions $T_1^*,\dots,T_t^*$ with replacement from $T_1,\dots,T_t$, then we can formally define a bootstrap sample of $\Mse_t-\mse_{\infty}$ according to
\begin{equation}\label{eqn:bootdef0}
\Mse_t^*-\Mse_t \ := \ \psi(\bar T_t^*)-\psi(\bar T_t),
\end{equation}
where $\bar T_t^*:=\frac{1}{t}\sum_{i=1}^t T_i^*$. In turn, after generating a collection of such bootstrap samples, we can use their empirical $(1-\alpha)$-quantile as an  estimate of $q_{1-\alpha}(t)$. However, as a technical point, it should be noted that~\eqref{eqn:bootdef0} is a ``theoretical'' bootstrap sample of $\Mse_t-\mse_{\infty}$, because the functional $\psi$ depends on the unknown distribution of the test point $\L(X,Y)$. Nevertheless, the same reasoning can still be applied by replacing $\psi$ with an estimate $\hat\psi$, which will be explained in detail later in this subsection. Altogether, the method is summarized by the following algorithm.
	\begin{algorithm}[H]
		\caption{Bootstrap method for estimating $q_{1-\alpha}(t)$}\label{alg:basic}
		\normalsize
		\begin{flushleft}
	\textbf{For} $b=1,\dots,B$\,\textbf{:}
	\begin{itemize}
	\item Sample $t$ functions $(T_1^*,\dots,T_t^*)$ with replacement from $(T_1,\dots,T_t)$.\\[.2cm]
	\item Compute the bootstrap sample $z_{t,b}:=\hat\psi(\bar T_t^*)-\hat\psi(\bar T_t)$.
	\end{itemize}	
	\textbf{Return:}\smash{ the empirical $(1-\alpha)$-quantile of $z_{t,1},\dots,z_{t,B}$ to estimate $q_{1-\alpha}(t)$.}
	\end{flushleft}
 	\end{algorithm}

\vspace{-0.3cm}

\paragraph{Using hold-out or out-of-bag samples.}  To complete our discussion of Algorithm~\ref{alg:basic}, it remains to clarify how the functional $\psi$ can be estimated from either hold-out samples, or so-called ``out-of-bag'' (\textsc{oob}) samples. With regard to the first case, suppose a set of $m$ labeled samples $\tilde \D=\{(\tilde X_1,\tilde Y_1),\dots,(\tilde X_m,\tilde Y_m)\}$ has been held out from the training set $\D$. Using this set, the estimate $\hat\psi(\bar T_t)$ in Algorithm~\ref{alg:basic} can be easily obtained as
\begin{equation}\label{eqn:hold1}
 \hat\psi(\bar T_t) \, = \, \frac{1}{m}\sum_{j=1}^m (\tilde Y_j-\bar T_t(\tilde X_j))^2.
\end{equation}
Analogously, we may also obtain $\hat\psi(\bar T_t^*)$ by using $\bar T_t^*$ instead of $\bar T_t$ in the formula above.

If the regression functions $T_1,\dots,T_t$ are trained via bagging or random forests, it is possible to avoid the use of a hold-out set by taking advantage of \textsc{oob} samples, which are a unique attribute of these methods. To define the notion of an \oob~sample, recall that these methods train each function $T_i$ using a random set $\D_i^*$ obtained from $\D$ by sampling with replacement. Due to this sampling mechanism, it follows that each set $\D_i^*$ is likely to exclude approximately  $(1-\frac{1}{n})^n\approx 37\%$ of the training points in $\D$. So, as a matter of terminology, if a particular training point $X_j$ does not appear in $\D_i^*$, we say that $X_j$ is ``out-of-bag'' for the function $T_i$. Also, we write $\oob(X_j)\subset\{1,\dots,t\}$ to denote the index set corresponding to the functions for which $X_j$ is \oob. 

From a statistical point of view, \oob~samples are important because they serve as ``effective'' hold-out points. (That is, if $X_j$ is \oob~for $T_i$, then the function $T_i$ ``never touched'' the point $X_j$ during the training process.) Hence, it is natural to consider the following alternative estimate of $\psi$ based on \oob~samples,
\begin{equation}
\hat\psi_{\textsc{o}}(\bar T_t) \, = \, \frac{1}{n}\sum_{j=1}^n (Y_j-\bar T_{t,\textsc{o}\!}(X_j))^2,
\end{equation}
where we define $\bar T_{t,\textsc{o}}(X_j)$ to be the average over the functions for which $X_j$ is \oob, 
$$\bar T_{t,\textsc{o}}(X_j)=\ts\frac{1}{|\textsc{oob}(X_j)|}\displaystyle\sum_{i\in\oob(X_j)} T_i(X_j),$$
and $|\cdot |$ refers to the cardinality of a set. Similarly, we define $\hat\psi_{\textsc{o}}(\bar T_t^*)$ by replacing each function $T_i$ above with $T_i^*$. Lastly, in the case when $\oob(X_j)$ is empty, we arbitrarily define $\bar T_{t,\textsc{o}}(X_j)=Y_j$, but this occurs very rarely. In fact, it can be checked that for a given point $X_j$, the set $\oob(X_j)$ is empty with probability approximately equal to $(0.63)^t$.
\vspace{-.2cm}
\subsection{Measuring convergence with respect to variable importance}\label{sec:VI}
\vspace{-.2cm}
In addition to their broad application in prediction problems, randomized ensembles have been very popular for the task of variable selection~\citep[e.g.][]{rf_gene_2006,strobl2008,ishwaranVI2007,genuer2010,louppe2013,genuer2015,gregorutti2017}. Although a variety of procedures have been proposed for variable selection in this context, they are generally based on a common approach of ranking the variables according to a measure of averaged variable importance (VI). Under this approach, the averaged VI assigned to each variable typically converges to a limiting value as the ensemble becomes large. However, in practice, the user does not know how this convergence depends on the ensemble size --- much like we have seen already for $\Mse_t$.

\paragraph{Uniform convergence across variables.} Before moving on to the details of our extended method, it is worth emphasizing an extra subtlety of measuring algorithmic convergence for VI. Specifically, we must keep in mind that because variable selection is based on ranking, it is important that algorithmic convergence is reached for \emph{all} variables. In other words, if the VI for some variables converges more slowly than for others, then the ranking of variables will be distorted by purely algorithmic effects. For this reason, our extended method will provide a way to ensure that algorithmic convergence is achieved uniformly across all variables.

\begin{figure}[H]
	\begin{minipage}{\textwidth}
		\begin{centering}
			
			\subfloat{
				\begin{overpic}[width=50mm,height=50mm,angle=270]{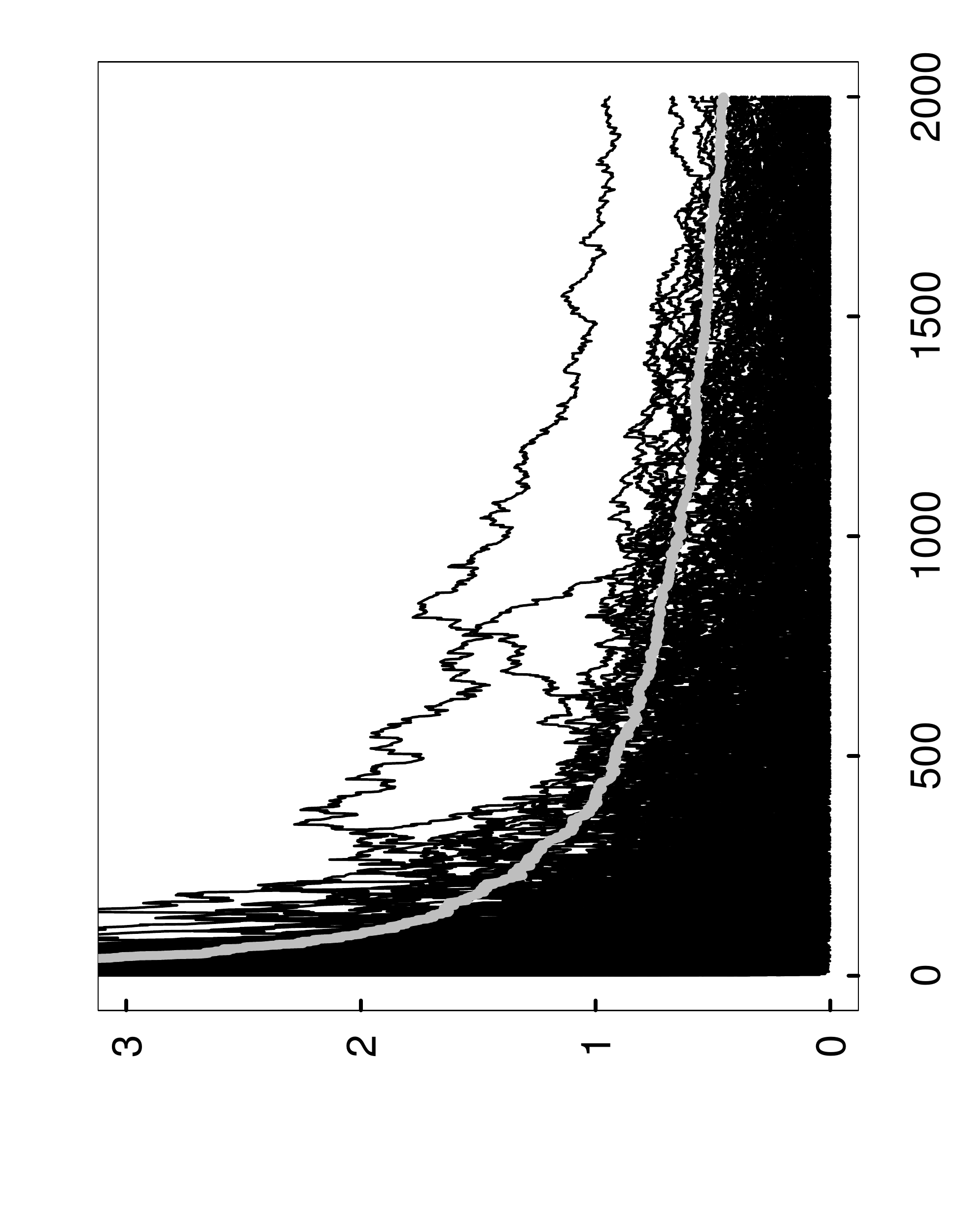} 
					\put(36,-5){\color{black}{\small ensemble size $t$}}   
					\put(2,27){\rotatebox{90}{\small $|\overline{\VI}_t(1)-\text{vi}_{\infty}(1)|$}}
					\put(52,65){\color{gray}{\Huge -}}
					\put(60,68){\color{black}{\footnotesize $0.9$ quantile}}
				\end{overpic}	
			}  \ \ \ \ \ \ \ \
			\subfloat{
				\begin{overpic}[width=50mm,height=50mm,angle=270]{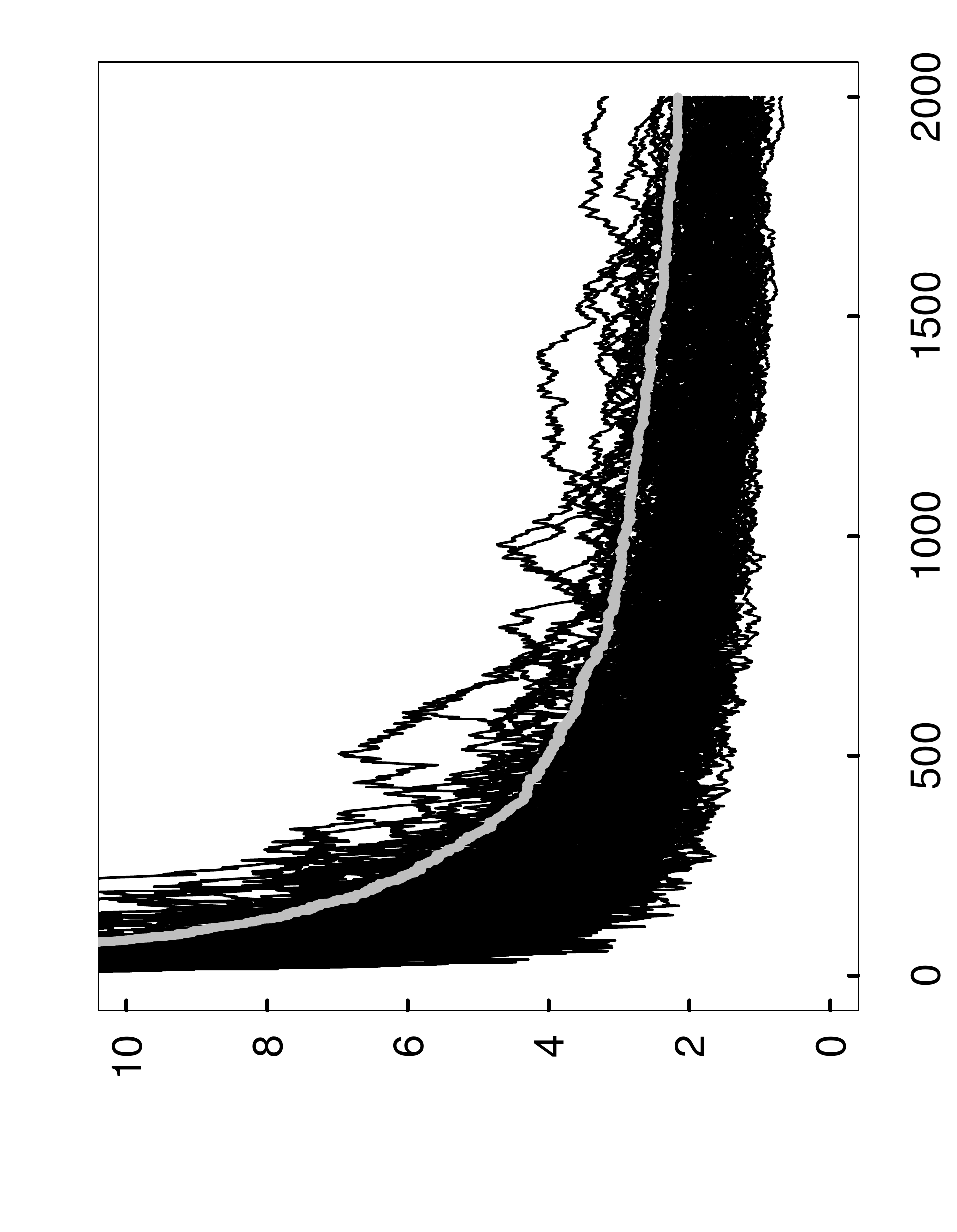} 
					\put(36,-5){\color{black}{\small ensemble size $t$}}   
					\put(-1,13){\rotatebox{90}{\small \ \ \ $ \displaystyle\max_{1\leq l\leq p}|\overline{\VI}_t(l)-\text{vi}_{\infty}(l)|$}}
					\put(52,65){\color{gray}{\Huge -}}
					\put(60,68){\color{black}{\footnotesize  $0.9$ quantile}}
				\end{overpic}	
			} 
			\par\end{centering}
		\vspace{0.2cm}
		\caption{Left Panel: 1,000 sample paths of $|\VI_t(1)-\VI_{\infty}(1)|$, with the true $0.9$ quantile curve in gray. Right Panel: 1,000 sample paths of the variable $\max_{1\leq l\leq p}|\VI_t(l)-\text{vi}_{\infty}(l)|$, with the true 0.9 quantile curve in gray. Both panels were obtained from the Music dataset described in Section~\ref{sec:expt}.
			\vspace{-0.0cm}
		}
		\label{fig:VI}
	\end{minipage}
\end{figure}
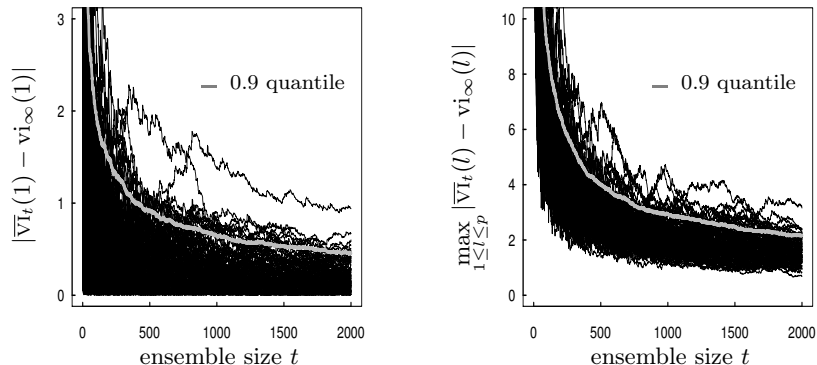

\paragraph{Setup for variable importance.} To describe algorithmic convergence for VI in detail, let $T_1,\dots,T_t$ be a randomized ensemble that satisfies the representation~\eqref{eqn:rep}, and consider a situation where the training samples have $p\geq 1$ variables (i.e.~the space $\mathcal{X}$ is $p$-dimensional).  Also, suppose that for each function $T_i$, we have a rule for assigning an importance value to each variable $l\in\{1,\dots,p\}$. Due to the fact that $T_i$ is a random function, it follows that the importance value is a random variable, denoted by  $\VI_i(l)$. (Choices for computing this will be discussed shortly.) Likewise, the vector of such values associated with $T_i$ is denoted $\VI_i=(\VI_i(1),\dots,\VI_i(p))$, and the averaged vector of importance measures is denoted as 
 \begin{equation}
  	\overline{\VI}_t=\frac{1}{t}\sum_{i=1}^t\VI_i.
 \end{equation}
Hence, by comparing the entries of the vector $\overline{\VI}_t=(\overline{\VI}_t(1),\dots,\overline{\VI}_t(p))$, the user is then able to rank the variables, and this is commonly done using a built-in option from the standard random forests software package~\citep{randomForestsCitation}. 

Up to this point, we have not specified a particular rule for computing the values $\VI_i(l)$, but several choices are available. For instance, two of the prevailing choices for regression are based on the notions of ``node impurity'' (for regression trees) or ``random permutations'' (for general regression functions).
However, from an abstract point of view, our proposed method does not depend on the underlying details of these rules, and so we refer to the book~\citep[][Sec 15.3.2]{ESL} for additional background. Indeed, our proposed method is applicable to any VI rule, provided that the random vectors $\VI_1,\dots,\VI_t$ are conditionally i.i.d.~given $\D$. In particular, this property is satisfied by both of the mentioned rules when $T_1,\dots,T_t$ follow the representation~\eqref{eqn:rep}.

When the conditional i.i.d.~property for $\VI_1,\dots,\VI_t$ holds and $\D$ is held fixed, the average $\overline{\VI}_t$ will generally converge to a limiting vector $\textup{vi}_{\infty}\in\R^p$ as $t\to\infty$. In order to measure this convergence uniformly across $l\in\{1,\dots,p\}$, we will focus on the (unknown) random variable
\begin{equation}
\ve_t:= \max_{1\leq l\leq p}|\overline{\VI}_t(l)-\textup{vi}_{\infty}(l)|,
\end{equation}
and our goal will be to estimate its $(1-\alpha)$-quantile, denoted as

\begin{equation}
\tq_{1-\alpha}(t) \, := \, \inf\Big\{q\in[0,\infty)\ \bigg|  \ \P\big(\ve_t\leq q\,\big| \D\big) \, \geq \, 1-\alpha\Big\}.
\end{equation}

\paragraph{The bootstrap method for variable importance.}  By analogy with our method for estimating the quantiles of $\Mse_t-\mse_{\infty}$, we propose to construct bootstrap samples of $\ve_t$ by resampling the vectors $\VI_1,\dots,\VI_t$, and then estimating $\tq_{1-\alpha}(t)$ with the empirical $(1-\alpha)$-quantile. In algorithmic form, the procedure works as follows.
	\begin{algorithm}[H]
		\caption{Bootstrap method for estimating $\tq_{1-\alpha}(t)$ }\label{alg:VI}
		\normalsize
		\begin{flushleft}
	\textbf{For} $b=1,\dots,B$\,\textbf{:}
	\begin{itemize}
	\item Sample $t$ vectors $(\VI_1^*,\dots,\VI_t^*)$ with replacement from $(\VI_1,\dots,\VI_t)$, and let $\overline{\VI}_t^*=\frac{1}{t}\sum_{i=1}^t\VI_i^*$.\\[.2cm]
	\item Compute the bootstrap sample $\ve_{t,b}^*:=\max_{1\leq l\leq p}|\overline{\VI}_t^*(l)-\overline{\VI}_t(l)|$.
	\end{itemize}	
	\textbf{Return:} the empirical $(1-\alpha)$-quantile of $\ve_{t,1}^*,\dots,\ve_{t,B}^*$ to estimate $\tq_{1-\alpha}(t)$.
	\end{flushleft}
 	\end{algorithm}

\vspace{-0.3cm}
~\\
Numerical results illustrating the performance of this algorithm, as well as Algorithm~\ref{alg:basic}, are given Section~\ref{sec:expt}.

\section{Main result}\label{sec:main}
In this section, we develop the main theoretical result of the paper, which guarantees that a bootstrap estimate of $q_{1-\alpha}(t)$ serves its intended purpose.
Namely, if this estimate is denoted by $\hat q_{1-\alpha}(t)$, then we will show that for a fixed set $\D$, the inequality
\begin{equation}
\Mse_t-\mse_{\infty} \, \leq \ \hat q_{1-\alpha}(t)
\end{equation}
holds with a probability that is effectively $1-\alpha$.

To establish this result, we will rely on a type of simplification that is commonly used in the analysis of bootstrap methods, which is to exclude sources of error beyond the resampling process itself. More specifically, we will focus 
on bootstrap samples of the form $\Mse_t^*-\Mse_t$ (defined in equation~\eqref{eqn:bootdef0}), since these are not affected by the extraneous error from estimating the functional $\psi$. A key benefit of this choice is that it clarifies how the performance of the bootstrap is related to the characteristics of the ensemble. Meanwhile, even with such a simplification, the proof of the result is still quite involved. Likewise, this choice was also used in our previous analysis of the classification setting for the same reasons~\citep{Lopes:2019}. Apart from this detail, the analysis in the current paper is entirely different.

With regard to the ensemble, the only assumptions used in our analysis are that it satisfies the representation~\eqref{eqn:rep}, as well as some basic moment conditions. From the standpoint of existing theory for randomized ensembles, these assumptions are very mild --- because the representation~\eqref{eqn:rep} is always satisfied by bagging and random forests. By contrast, it is much more common in the theoretical literature to work with ensembles that are simpler than the ones used in practice; and indeed, our previous work in the classification setting relied on a highly specialized type of ensemble. 
Furthermore, the moment parameters in our current result are guaranteed to be finite in the important case when $T_1,\dots,T_t$ are trained by CART, as will be explained shortly. Finally, it is notable that our result is fully~\emph{non-asymptotic}, whereas much existing work on the convergence of randomized ensembles has taken an asymptotic approach that does not always provide explicit rates of convergence.

\paragraph{Notation.} If $g$ and $h$ are real-valued functions on $\mathcal{X}\times \R$, we denote their inner product with respect to the test point distribution $\nu=\mathcal{L}(X,Y)$ as
$$\langle g,h\rangle =\int_{\mathcal{X}\times \R} g(x,y)\,h(x,y)\,d\nu(x,y),$$
and accordingly, we write $\|g\|_{L_2}=\sqrt{\langle g,g\rangle}$.
In addition, recall the function $\vartheta(x)=\E[T_1(x)|\D]$ from equation~\eqref{eqn:varthetadef0}, and define the random variable\label{momentpage}
\begin{equation}\label{eqn:zetaidef}
\zeta = 2\,\langle \vartheta -y, T_1-\vartheta\rangle,
\end{equation}
where the expression $\vartheta-y$ is interpreted as the function $(x,y)\mapsto \vartheta(x)-y$.
When the random variable $\zeta$ is conditioned on $\D$, we denote its standard deviation by
$$\sigma(\D)=\sqrt{\var(\zeta|\D)},$$
and the finiteness of this quantity will follow from assumption~\textbf{A2} below. Also, all expressions involving $1/\sigma(\D)$ will be understood as $\infty$ in the exceptional case when $\sigma(\D)=0$. Lastly, for each positive integer $k$, we define the moment parameter
$$\beta_k(\D)=\max\Big\{\big(\E\big[\|T_1-y\|_{L_2}^{2k}\big|\D\big]\big)^{1/k}\, ,\, \big(\E\big[ \|T_1-\vartheta\|_{L_2}^{2k}|\D\big]\big)^{1/k}\Big\},$$
which provides a convenient way to quantify the tail behavior of the random variables $\|T_1-y\|_{L_2}$ and $\|T_1-\vartheta\|_{L_2}$.

\paragraph{Assumptions.} With the above notation in place, we can state the two assumptions needed for our main result.
~\\

\noindent \textbf{A1.} The ensemble $T_1,\dots,T_t$ can be represented in the form~\eqref{eqn:rep}.\\

\noindent \textbf{A2.} There is at least one integer $k\geq 2$ such that $\beta_{3k}(\D)<\infty$.\\

\noindent To interpret these assumptions, recall that \textbf{A1} is always satisfied by bagging and random forests, as explained in Section~\ref{sec:setup}. Regarding the finiteness of $\beta_{3k}(\D)$ in \textbf{A2}, it is noteworthy that this condition is satisfied for \emph{arbitrarily large} values of $k$ whenever the functions $T_1,\dots,T_t$ are trained by the standard method of CART. This is because the range of the functions is determined by the training labels $Y_1,\dots, Y_n$. In particular, if we put $M(\D)\!:=\max_{1\leq i\leq n} |Y_i|$, then every tree $T_i$ satisfies the bound $\sup_{x\in\mathcal{X}} |T_i(x)| \ \leq \ M(\D)$, which implies
\begin{equation}\label{eqn:betabound}
\beta_k(\D)\leq 4M(\D)^2,
\end{equation}
for every $k$. The same reasoning also applies beyond CART to any other method whose predictions fall within the range of the training labels. We now state the main result of the paper.

\begin{theorem}\label{thm:main}
Suppose that \textbf{\textup{A1}} and \textbf{\textup{A2}} hold. In addition, let $k\geq2$ be as in \textbf{\textup{A2}}, and let  \ $\hat q_{1-\alpha}(t)$ denote the empirical $(1-\alpha)$-quantile of $B$ bootstrap samples of the form~\eqref{eqn:bootdef0}.
Lastly, define the quantity
 \begin{equation}\label{eqn:deltadef}  
 \delta_{t,k,B}(\D) \ := \ \ts\frac{k^2}{\sqrt{t}}\Big(\ts\frac{\beta_{3k}(\D)}{\sigma(\D)}\Big)^3 \ + \ e^{-k/2} \ + \ \sqrt{\frac{\log(B)}{B}}.
 \end{equation}
Then, there is an absolute constant $c_0>0$ such that $\hat{q}_{1-\alpha}(t)$ satisfies
\begin{equation}\label{eqn:thmresult}
\P\Big(\Mse_t-\textup{mse}_{\infty}\leq \,\hat q_{1-\alpha}(t) \,\Big |\, \D\Big) \ \geq \ 1-\alpha-c_0\,\delta_{t,k,B}(\D).
\end{equation}

\end{theorem}

\paragraph{Remarks.} In essence, the result shows that $\hat q_{1-\alpha}(t)$ bounds the unknown convergence gap $\Mse_t-\mse_{\infty}$ with a probability that is not much less than the ideal value of $1-\alpha$. To comment on some further aspects of the result, note that the inequality~\eqref{eqn:thmresult} has the desirable property of being \emph{scale-invariant} with respect to the labels $Y_1,\dots, Y_n$ and the functions $T_1,\dots, T_t$. More precisely, if we were to change the units of the labels and functions by a scale factor $c>0$, it can be checked that 
both sides of ~\eqref{eqn:thmresult} would remain unchanged.

Another important aspect of Theorem~\ref{thm:main} deals with the dependence of $\delta_{t,k,B}(\D)$ on the value of $k$. Specifically, it is interesting to develop a bound on $\delta_{t,k,B}(\D)$ that simplifies the role of $k$. To do this, we now consider the situation when the regression functions are trained by CART, or more generally, when the boundedness condition~$\beta_k(\D)\leq 4M(\D)^2$ holds for every $k\geq 1$, as in~\eqref{eqn:betabound}. In such cases,
 we may evaluate the particular choice 
\begin{equation}\label{eqn:kchoice}
k\,=\lceil \log(t)-4\log\log(t)\rceil,
\end{equation}
 which leads to the following bounds,
$$e^{-k/2} \ \leq \ \frac{\log(t)^2}{\sqrt{t}} \ \ \ \ \text{ and } \ \ \ \ \frac{k^2}{\sqrt{t}} \ \leq \ \frac{c_0\log(t)^2}{\sqrt{t}},$$
 for some absolute constant $c_0>0$ and all $t\geq 2$.
These bounds imply that there is a number $c(\D)>0$ not depending on $t$, $k$, or $B$,  such that 
\begin{equation}
 \delta_{t,k,B}(\D) \, \ \leq \ \, \frac{c(\D)\, \log(t)^2}{\sqrt{t}} \, \ + \ \,\sqrt{\frac{\log(B)}{B}},
\end{equation}
which considerably simplifies the interpretation of $\delta_{t,k,B}(\D)$. Hence, at a high level, this indicates that as long as the regression functions have well-behaved moments, then for a fixed set $\D$, the quantity $\delta_{t,k,B}(\D)$ converges to 0 at \emph{nearly parametric rates} with respect to both $t$ and $B$.

\section{Computation and speedups}\label{sec:comp}

In order for the proposed method to be a practical a tool for checking algorithmic convergence, its computational cost should be manageable in comparison to training the ensemble itself. Below, in Section~\ref{sec:cost}, we offer a quantitative comparison, showing that under simple conditions, Algorithms~\ref{alg:basic} and~\ref{alg:VI} are not a bottleneck in relation to training $t$ regression functions with CART. Additionally, we show in Section~\ref{sec:extrap} how an extrapolation technique from our previous work on classification can be improved in our current setting with a \smash{\emph{bias correction}} \emph{rule}.

\subsection{Cost comparison}\label{sec:cost}
Because the CART method is based on a greedy iterative algorithm, the exact computational cost of training a regression tree is difficult to describe. For this reason, the authors of CART analyzed its cost in the simplified situation where each node of a regression tree is split into exactly 2 child nodes (except for the leaves). To be more precise, suppose $\mathcal{X}\subset\R^p$, and let $d\geq 2$ denote the ``depth'' of the tree, so that there are $2^d$ leaves. In addition, suppose that when the algorithm splits a given node, it searches over $\lceil p/3\rceil$ candidate variables that are randomly chosen from $\{1,\dots,p\}$, which is the default rule when CART is used by random forests~\citep{randomForestsCitation}. Based on these assumptions, the analysis in the book~\citep[][p.166]{CART} shows that the number of operations involved in training $t$ such trees is at least of order $\Omega(t\cdot p\cdot d\cdot n)$.

\paragraph{The cost of Algorithm~\ref{alg:basic}.} To determine the cost of Algorithm~\ref{alg:basic}, it is important to clarify that when bagging and random forests are used in practice, the prediction error of the ensemble is typically estimated automatically using either hold-out or \oob~samples.
As a result, the predicted values of each tree on these samples can be regarded as being pre-computed by the ensemble method. Once these values are available, the subsequent cost of Algorithm~\ref{alg:basic} is simple to measure. Specifically, in the case of hold-out samples, equation~\eqref{eqn:hold1} shows that the cost to obtain $\hat\psi(\bar T_t)-\hat\psi(\bar T_t^*)$ for each bootstrap sample is $\mathcal{O}(t\cdot m)$, which leads to an overall cost that is $\mathcal{O}(B\cdot t\cdot m)$. Similarly, for the case of~\oob~samples, the overall cost is $\mathcal{O}(B\cdot t\cdot n)$. Altogether, this leads to the conclusion that the cost of Algorithm~\ref{alg:basic} does not exceed that of training the ensemble if the number of bootstrap samples satisfies the very mild condition
\begin{equation}\label{eqn:Bcond1}
B=\mathcal{O}(p\cdot d),
\end{equation}
and this applies to either the hold-out or \oob~cases, provided $m=\mathcal{O}(n)$. Moreover, our discussion in Section~\ref{sec:extrap} will show that the condition~\eqref{eqn:Bcond1} can even be further relaxed via extrapolation.

Beyond the fact that Algorithm~\ref{alg:basic} compares well with the cost of training an ensemble, there are several other favorable aspects worth mentioning. First, the algorithm only relies on predicted labels for its input, and it never needs to access any points in the space~$\mathcal{X}$. In particular, this means that the cost of the algorithm is \emph{independent of the dimension of} $\mathcal{X}$. Second,  the bootstrap samples in Algorithm~\ref{alg:basic} are simple to compute in parallel, which means that the cost of the algorithm can be reduced approximately by a factor of $B$. 

\paragraph{The cost of Algorithm~\ref{alg:VI}.} Many of the previous considerations for Algorithm~\ref{alg:basic} also apply to Algorithm~\ref{alg:VI}, but it turns out that the cost of Algorithm~\ref{alg:VI} can be much  less when $n$ is large. Because each bootstrap sample  in Algorithm~\ref{alg:VI} requires forming an average of $t$ vectors in $\R^p$, it is straightforward to check that the overall cost is $\mathcal{O}(B\cdot t\cdot p)$, where we view the vectors $\VI_1,\dots,\VI_t$ as being pre-computed by the ensemble method. In particular, it is worth emphasizing that the cost of the algorithm is \emph{independent of $n$}, and is thus highly scalable. Furthermore, under the setup of our earlier cost comparison with CART, the cost of Algorithm~\ref{alg:VI} does not exceed the cost of training the ensemble if
$$B=\mathcal{O}(n\cdot d),$$
which allows for plenty of bootstrap samples in practice. In fact, our numerical experiments show that even $B=50$ can work well when $n$ is on the order of $10^4$, indicating that Algorithm~\ref{alg:VI} is quite inexpensive in comparison to training.

\subsection{Further reduction of cost by extrapolation}\label{sec:extrap}
The basic idea of extrapolation is to check algorithmic convergence for a small ``initial'' ensemble, say of size $t_0$, and then use this information to ``look ahead'' and predict convergence for a larger ensemble of size $t> t_0$.  
This general technique has a long history in the development of numerical algorithms, and further background can be found in~\citep{bickelrichardson,brezinski,sidi} as well as references therein. In the remainder of this section, we first summarize how extrapolation was previously developed in our companion paper~\citep{Lopes:2019}, and then explain how that approach can be improved with a bias correction for \oob~samples.

\paragraph{A basic version of extrapolation.} At a technical level, our use of extrapolation is based on the central limit theorem, which suggests that the fluctuations of $\Mse_t-\mse_{\infty}$ should scale like $1/\sqrt{t}$ as a function of $t$. As a result, we expect that the quantile $q_{1-\alpha}(t)$ should behave like 
$$q_{1-\alpha}(t)\approx \ts\frac{\kappa}{\sqrt{t}},$$
 for some quantity $\kappa$ that may depend on all problem parameters except $t$.

To take advantage of this heuristic scaling property, suppose that we train an initial ensemble of size $t_0$, and run Algorithm~\ref{alg:basic} to obtain an estimate $\hat q_{1-\alpha}(t_0)$. We can then extract an estimate of $\kappa$ by defining
$$\hat\kappa:= \sqrt{t_0} \,\hat q_{1-\alpha}(t_0).$$
Next, we can rapidly estimate $q_{1-\alpha}(t)$ for all subsequent $t\geq t_0$ by defining the extrapolated estimate
\begin{equation}\label{eqn:extrapdef0}
\hat q_{1-\alpha}^{\text{\ ext}}(t)  \, := \, \ts\frac{\hat\kappa}{\sqrt{t}} \, = \, \ts\frac{\sqrt{t_0} \hat q_{1-\alpha}(t_0)}{\sqrt {t}}.
\end{equation}
In particular, there are two crucial benefits of this estimate: (1) It is much faster to apply Algorithm~\ref{alg:basic} to a small initial ensemble of size $t_0$ than to a large one of size $t$. (2) If we would like $\Mse_t$ to be within some tolerance $\e>0$ of the limit $\mse_{\infty}$, then we can use the condition 
$$\hat q_{1-\alpha}^{\ \text{ext}}(t)\leq \e$$ 
to \emph{dynamically predict} how large $t$ must be chosen to reach that tolerance, namely
$t\geq (\sqrt{t_0}\hat q_{1-\alpha}(t_0)/\e)^2$.

\paragraph{Bias-corrected extrapolation.} If the initial estimate $\hat q_{1-\alpha}(t_0)$ is obtained by implementing Algorithm~\ref{alg:basic} with \oob~samples, it turns out to be a biased estimate of $q_{1-\alpha}(t_0)$. Fortunately however, it is possible to correct for this bias in a simple way, as we now explain.

To understand the source of the bias, recall that for each point $X_j$ in the training set, we write $\oob(X_j)\subset\{1,\dots,t\}$ to index the regression functions for which $X_j$ is \oob. Also, it is simple to check that for an initial ensemble of size $t_0$, the expected cardinality of $\oob(X_j)$ is given by
\begin{equation}\label{eqn:tau0def}
\tau_n(t_0):=(1-1/n)^n \cdot t_0.
\end{equation}
In other words, this means that when an ensemble of size $t_0$ makes a prediction on an \oob~point, the ``effective'' size of the ensemble is $\tau_n(t_0)$, rather than $t_0$. As a result, if we implement Algorithm~\ref{alg:basic} using \oob~samples with an initial ensemble of size $t_0$, then the output $\hat q_{1-\alpha}(t_0)$ should really be viewed as an estimate of $q_{1-\alpha}(\tau_n(t_0))$, rather than $q_{1-\alpha}(t_0)$. 

Based on this reasoning, we can adjust our previous definition of the estimate $\hat q_{1-\alpha}^{\ \text{ext}}(t)$ in~\eqref{eqn:extrapdef0} by using
\begin{equation}\label{eqn:extrapdef1}
\ts\hat q_{1-\alpha}^{\text{\ ext},\textsc{o}}(t) :=\ts\frac{\sqrt{\tau_n(t_0)}\hat q_{1-\alpha}(t_0) }{\sqrt{t}} \ \ \ \ \ \text{ for } \ \ \ \ \ t\geq \tau_n(t_0).
\end{equation}
Later on, in Section~\ref{sec:expt} we will demonstrate that this simple adjustment works quite well in practice. 

\paragraph{Remark.} As a clarification, it should be noted that the definition~\eqref{eqn:extrapdef1} is only to be used when Algorithm~\ref{alg:basic} is implemented with \oob~samples, and the basic rule~\eqref{eqn:extrapdef0} should be used in the case of hold-out samples. Also, the basic rule~\eqref{eqn:extrapdef0} can be easily adapted to extrapolate the estimate produced by Algorithm~\ref{alg:VI}, and so we omit the details in the interest of brevity.

\section{Numerical results}\label{sec:expt}
We now demonstrate the bootstrap's numerical accuracy at the tasks of measuring algorithmic convergence with respect to both mean-squared error and variable importance.
Overall, our results show that the extrapolated \oob~estimate is accurate at predicting the effect of increasing $t$. In fact, the results show that extrapolation succeeds at predicting what will happen when $t$ is increased by a factor of 4 beyond $t_0$, and possibly much farther.

\subsection{Organization of experiments } 

\paragraph{Data preparation.} Our experiments were based on several natural datasets that were each randomly partitioned in the following way. Letting $\mathcal{F}$ denote the full set of observation pairs $(X_1,Y_1),(X_2,Y_2),\dots$  for a given dataset, we evenly split $\mathcal{F}$ into a disjoint union $\mathcal{F}=\mathcal{D}\sqcup \mathcal{T}$, where the set $\mathcal{D}$ was used for training, and the set $\mathcal{T}$ was used to approximate the true quantile curves (namely~$q_{1-\alpha}(t)$ or $\tq_{1-\alpha}(t)$) for assessing algorithmic convergence. 

Since Algorithm~\ref{alg:basic} relies on a hold-out set, we also used a relatively small subset $\mathcal{H}\subset\mathcal{T}$ for that purpose. Specifically, the hold-out set $\mathcal{H}$ was chosen so that its cardinality satisfied $|\mathcal{H}|/(|\mathcal{H}|+|\mathcal{D}|)\approx 1/6$. This reflects a practical situation where the user can only afford to allocate $1/6$ of the available data for the hold-out set. In other words, the idea is to think of the user as only having access to $\mathcal{D}\sqcup\mathcal{H}$, with the set $\mathcal{T}$ as being used externally to establish ``ground truth'' for the rate of algorithmic convergence.

Each of the full datasets are briefly summarized below.
\begin{itemize}

\item \emph{Diamond}: This dataset is available in the package~{\tt{ggplot2}}~\citep{GGplot2}, and has been downsampled to 10,000 observations.
Each observation contains 9 measured features of a distinct diamond, and the features are used to predict the diamond's price. \\

\item \emph{Housing}: This dataset originates from 1990 California census and is available as part of the online supplement to the book~\citep{geron2017}. The observations are correspond to 20,640 homes, and for each home there are 9 features for predicting the home's price.\\

\item \emph{Music}: This dataset consists of 1,059 audio recordings (observations) described by 68 features that are used to predict the geographic latitude of the recording, as described in~\citep{zhou2014predicting}. The dataset is available at the UCI repository~\citep{Dua:2019} under the title \emph{Geographical Origin of Music Data Set}. \\

\item \emph{Protein}: This is dataset was collected from the fifth through ninth series of CASP experiments~\citep{CASP9}, and is available at the UCI repository~\citep{Dua:2019} under the title \emph{Physicochemical Properties of Protein Tertiary Structure Data Set}. The 45,730 observations correspond to artificially generated conformations of proteins (known as decoys) that are described by 9 biophysical features. Each decoy can be thought of as a perturbation of an associated ``target'' protein, and the features are used to predict how far the decoy is from its target.

\end{itemize}

\paragraph{Computing the true quantile curves $q_{1-\alpha}(t)$ and $\tq_{1-\alpha}(t)$.} Once a full dataset $\mathcal{F}$ was partitioned as above, we ran the random forests algorithm 1,000 times on the associated set $\mathcal{D}$, using the {\tt{R}} package {\tt{randomForest}}~\citep{randomForestsCitation}. The overall process was a serious computational undertaking, because $2,\!000$ regression trees were trained during every run, and hence a total of  $1,\!000\times 2,\!000=2\times 10^6$ trees were trained on each dataset.

During each run, as the ensemble size increased from $t=1$ to $t=2,\!000$, the corresponding true values of $\Mse_t$ were approximated with the ensemble's error rate on $\mathcal{T}$. Also, the true value of $\mse_{\infty}$ was approximated with the average of the 1,000 realizations of $\Mse_{2,000}$. In this way, the collection of runs produced 1,000 approximate sample paths of $\Mse_t-\mse_{\infty}$, similar to those illustrated in the right panel of Figure~\ref{fig:raw-data}. Finally, the quantile curve $q_{.90}(t)$ was extracted by using the empirical 90\% quantile of the 1,000 values of $\Mse_t-\mse_{\infty}$ at each $t=1,\dots,2,\!000$.

To handle the setting of variable importance, essentially the same steps were used. Specifically, we computed the vector $\overline{\VI}_t\in\R^p$ at every value $t=1,\dots,2,\!000$, for each of the 1,000 runs mentioned above. In addition, we approximated the vector $\text{vi}_{\infty}\in\R^p$ with the average of the 1,000 realizations of $\overline{\VI}_{2,000}$. Altogether, these computations provided us with 1,000 approximate sample paths of $\ve_t=\max_{1\leq l\leq p}|\overline{\VI}_t(l)-\text{vi}_{\infty}(l)|$, and then we used the empirical 90\% quantile at each $t=1,\dots,2,\!000$ to approximate $\tq_{.90}(t)$.

\paragraph{Applying the bootstrap algorithms with extrapolation.} For each of the described 1,000 runs of random forests, we applied the extrapolated versions of Algorithms~\ref{alg:basic} and~\ref{alg:VI} at the initial ensemble size of $t_0=500$, using a small number of $B=50$ bootstrap samples. Hence, this provided us 1,000 realizations of each type of the proposed estimates, which allows for an assessment of their variability. 

Below, in Sections~\ref{sec:nummse} and~\ref{sec:numVI}, we will show the results obtained by extrapolating to the final ensemble size of $t=2,\!000$.
In addition, for Algorithm~\ref{alg:basic}, we implemented both of the hold-out and \textsc{oob} versions, including the bias correction for the \oob~samples described in equation~\eqref{eqn:extrapdef1}.

\subsection{Numerical results for mean-squared error}\label{sec:nummse}

\begin{figure}[htbp]
	\centering
	\begin{minipage}[t]{0.48\textwidth}
		\centering
		\begin{overpic}[width=50mm,height=50mm,angle=270]{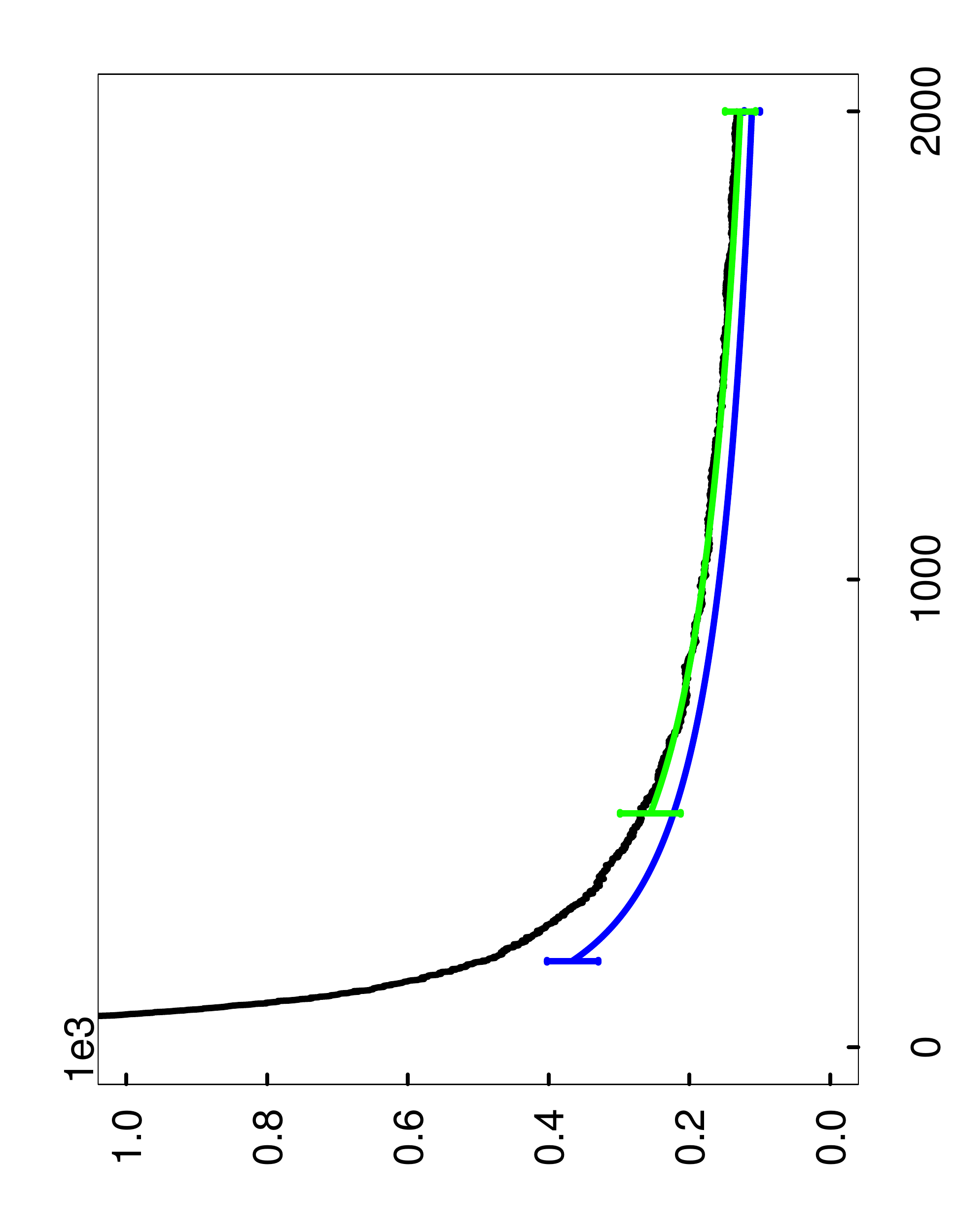} 
			\put(36,-5){\color{black}{\small ensemble size $t$}}   
			\put(-8,45){\rotatebox{90}{\small $q_{0.9}(t)$}}
			\put(40,65){\color{black}{\Huge -}}
			\put(48,68){\color{black}{\small $q_{0.9}(t)$ (true)}}
			\put(40,58){\color{blue}{\Huge -}}
			\put(48,61){\color{black}{\small \oob\ }}
			\put(40,51){\color{green}{\Huge -}}
			\put(48,54){\color{black}{\small hold-out}}
		\end{overpic}
		\vspace{0.2cm}
		\caption{Housing Data}\label{fig:housing}
	\end{minipage}
	\begin{minipage}[t]{0.48\textwidth}
		\centering
		\begin{overpic}[width=50mm,height=50mm,angle=270]{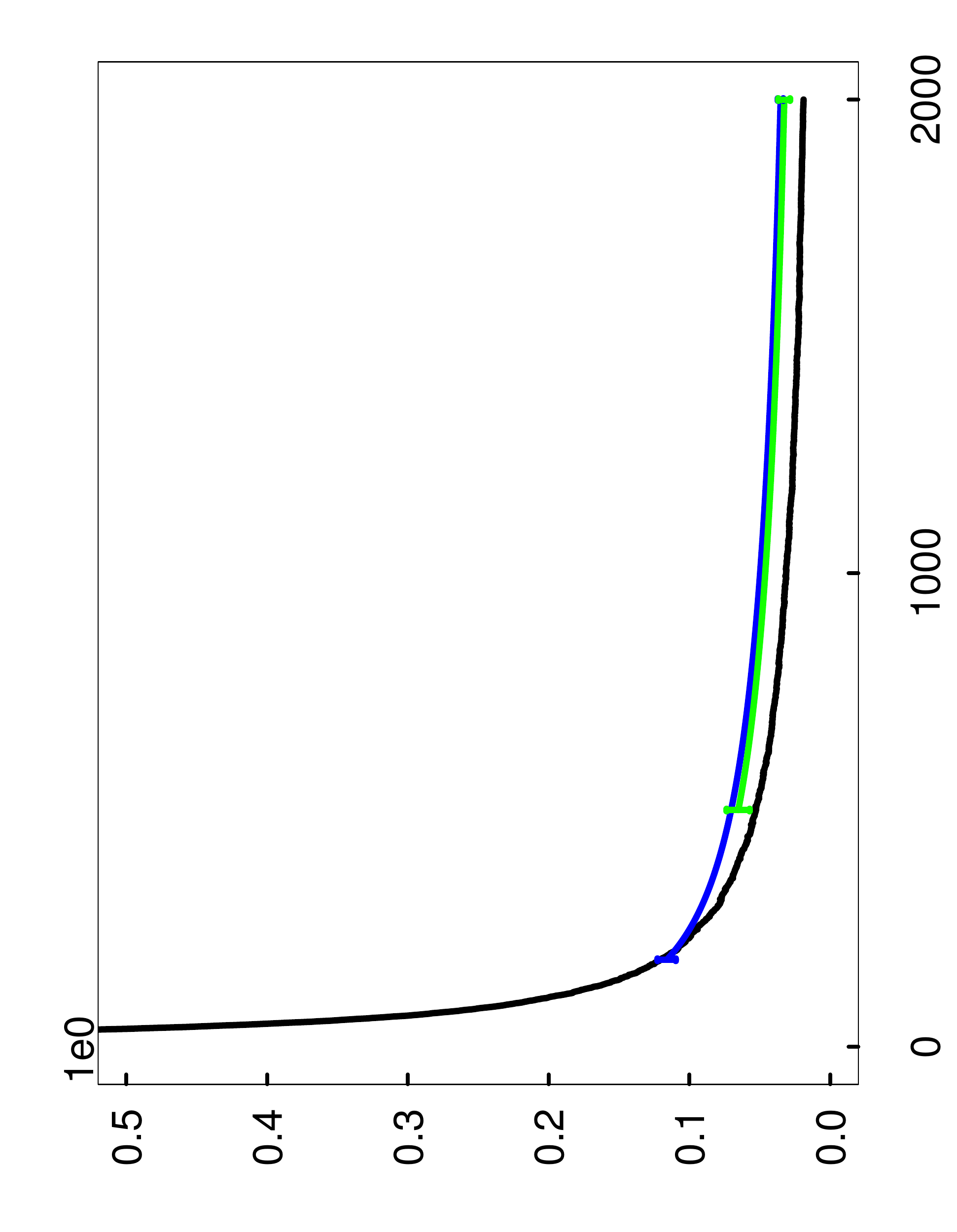} 
			\put(36,-5){\color{black}{\small ensemble size $t$}}   
			\put(-8,45){\rotatebox{90}{\small $q_{0.9}(t)$}}
			\put(40,65){\color{black}{\Huge -}}
			\put(48,68){\color{black}{\small $q_{0.9}(t)$ (true)}}
			\put(40,58){\color{blue}{\Huge -}}
			\put(48,61){\color{black}{\small \oob\ }}
			\put(40,51){\color{green}{\Huge -}}
			\put(48,54){\color{black}{\small hold-out}}
		\end{overpic}
				\vspace{0.2cm}
		\caption{Protein Data}\label{fig:protein}
	\end{minipage}
\end{figure}

\begin{figure}[htbp]
	\centering
	\begin{minipage}[t]{0.48\textwidth}
		\centering
		\begin{overpic}[width=50mm,height=50mm,angle=270]{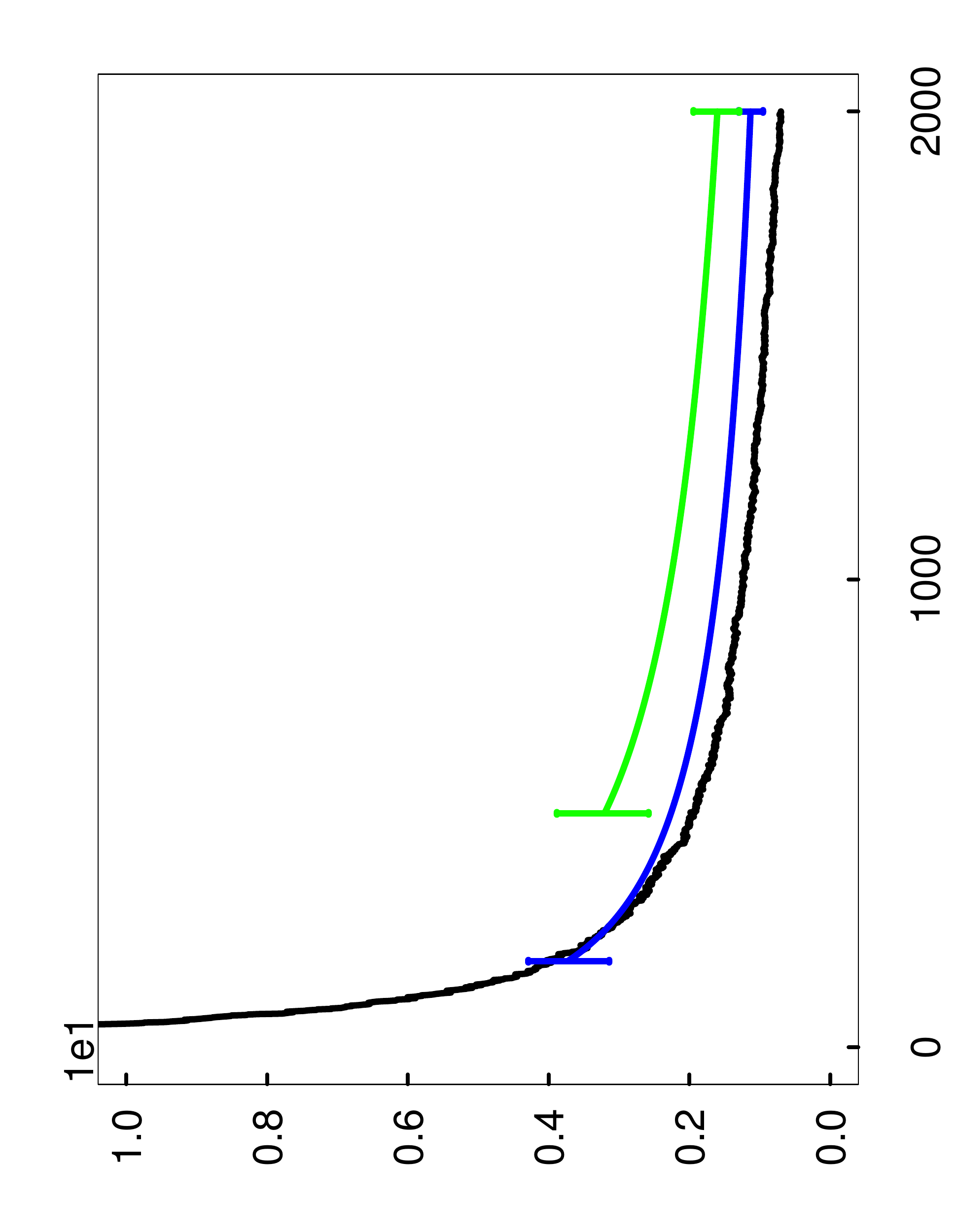} 
			\put(36,-5){\color{black}{\small ensemble size $t$}}   
			\put(-8,45){\rotatebox{90}{\small $q_{0.9}(t)$}}
			\put(40,65){\color{black}{\Huge -}}
			\put(48,68){\color{black}{\small $q_{0.9}(t)$ (true)}}
			\put(40,58){\color{blue}{\Huge -}}
			\put(48,61){\color{black}{\small \oob }}
			\put(40,51){\color{green}{\Huge -}}
			\put(48,54){\color{black}{\small hold-out}}
		\end{overpic}
				\vspace{0.2cm}
		\caption{Music Data}\label{fig:music}
	\end{minipage}
	\begin{minipage}[t]{0.48\textwidth}
		\centering
		\begin{overpic}[width=50mm,height=50mm,angle=270]{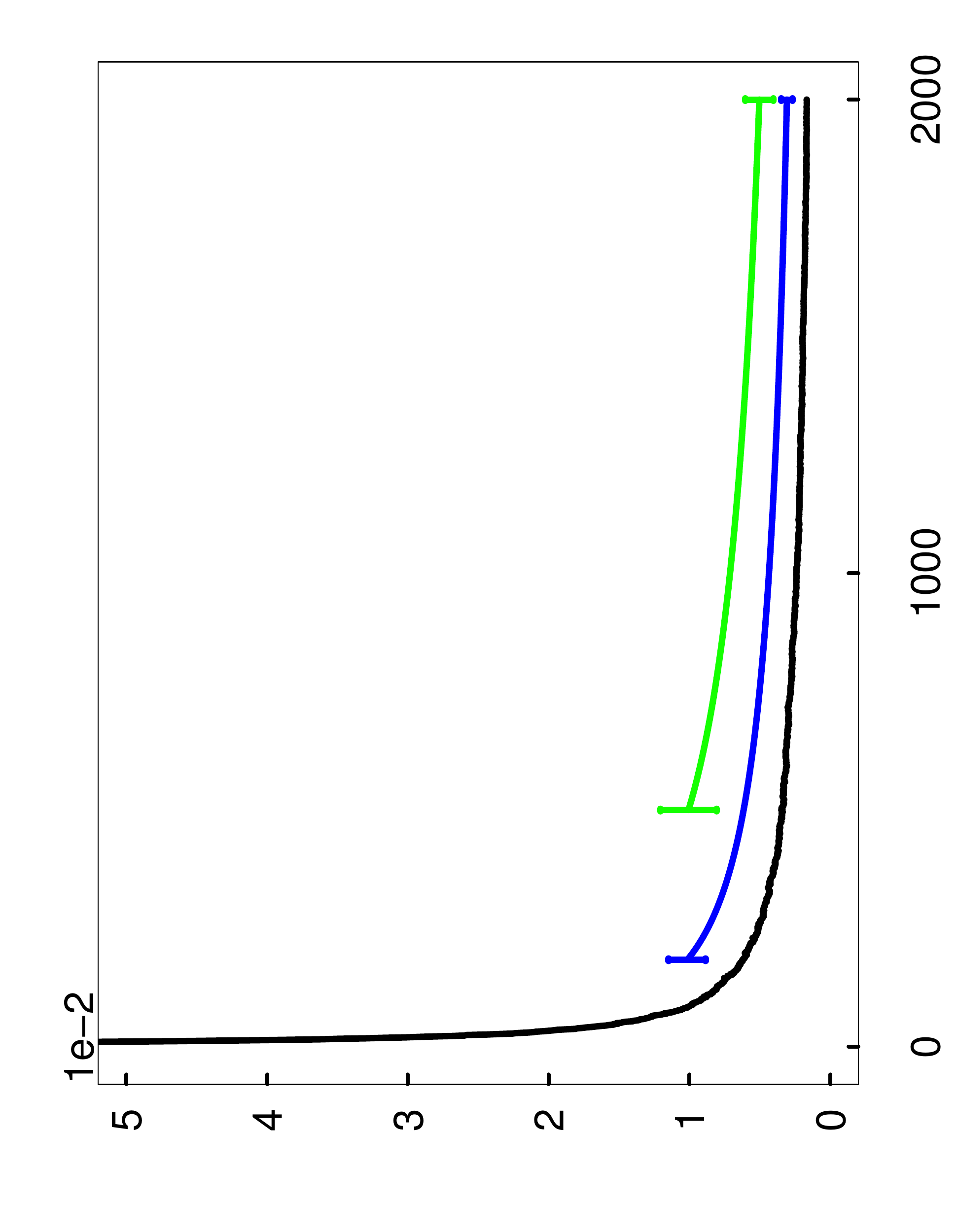} 
			\put(36,-5){\color{black}{\small ensemble size $t$}}   
			\put(-8,45){\rotatebox{90}{\small $q_{0.9}(t)$}}
			\put(40,65){\color{black}{\Huge -}}
			\put(48,68){\color{black}{\small $q_{0.9}(t)$ (true)}}
			\put(40,58){\color{blue}{\Huge -}}
			\put(48,61){\color{black}{\small \oob }}
			\put(40,51){\color{green}{\Huge -}}
			\put(48,54){\color{black}{\small hold-out}}
		\end{overpic}
				\vspace{0.2cm}
		\caption{Diamond Data}\label{fig:diamond}
	\end{minipage}
\end{figure}

 \paragraph{Organization of the plots.} The two types of estimates for $q_{.90}(t)$ are illustrated in Figures~\ref{fig:housing} through~\ref{fig:diamond}, with the hold-out estimator in green, and the \oob~estimator in blue. More specifically, these curves represent the averages of the estimates over the 1,000 runs described above, and the error bars display the fluctuations of the estimates over repeated runs ---corresponding to the 10th and 90th percentiles of the estimates. For the values of $t$ between the endpoints, we omit the error bars for clarity. Also, it is important to emphasize that these error bars should \emph{not} be interpreted as confidence intervals for $q_{.90}(t)$, and are only intended to show that the estimates have low variance.

  With regard to computation, another point to mention is that the estimates were only computed for the initial ensemble size $t_0=500$, and the rest of the green and blue curves were obtained essentially \emph{for free} by extrapolation. Lastly, as a clarification, it should be noted that the blue \oob~curve is shifted to the left of the green hold-out curve because of the bias correction rule~\eqref{eqn:extrapdef1} for \oob~samples.

\paragraph{Remarks on performance.} The main point to take away from the plots is that the \oob~estimate performs quite well overall, and can be much more accurate than the hold-out estimate (cf. Figures~\ref{fig:music} and~\ref{fig:diamond}). Furthermore, the \oob~estimate has an extra advantage because it does not require the user to hold out any data. For these reasons, we recommend the \oob~estimate in practice.

Another conclusion to draw from the plots is that the bias correction plays a significant role in the  extrapolation of the \oob~estimate. If the bias correction were not used, this would be equivalent to shifting the blue curve so that it starts at the same point as the green curve, which would clearly lead to a loss in accuracy. Also, it is remarkable that the extrapolated \oob~estimator
continues to be accurate at a final ensemble size of $t=2,\!000$ that is 4 times larger than the initial ensemble size $t_0=500$. Hence, this provides the user with a very inexpensive way to predict how quickly the ensemble will converge.  Moreover, even in the cases where the extrapolation starts from a mediocre initial estimate, the accuracy tends to improve as $t$ becomes larger.

To explain the inferior performance of the hold-out estimate, recall that it uses the small set $\mathcal{H}$ in order to estimate $\Mse_t$. As a result, the estimates of $\Mse_t$ using $\mathcal{H}$ have much more variability, which inflates the upper extremes of the estimator's sampling distribution, and thus leads to a larger estimate of $q_{.90}(t)$. On the other hand, the \oob~estimator is able to take advantage of the \oob~samples in the much larger set $\mathcal{D}$, which reduces this detrimental effect.


\newpage

\subsection{Numerical results for variable importance}\label{sec:numVI}
\begin{figure}[htbp]
	\centering
	\begin{minipage}[t]{0.48\textwidth}
		\centering
		\begin{overpic}[width=50mm,height=50mm,angle=270]{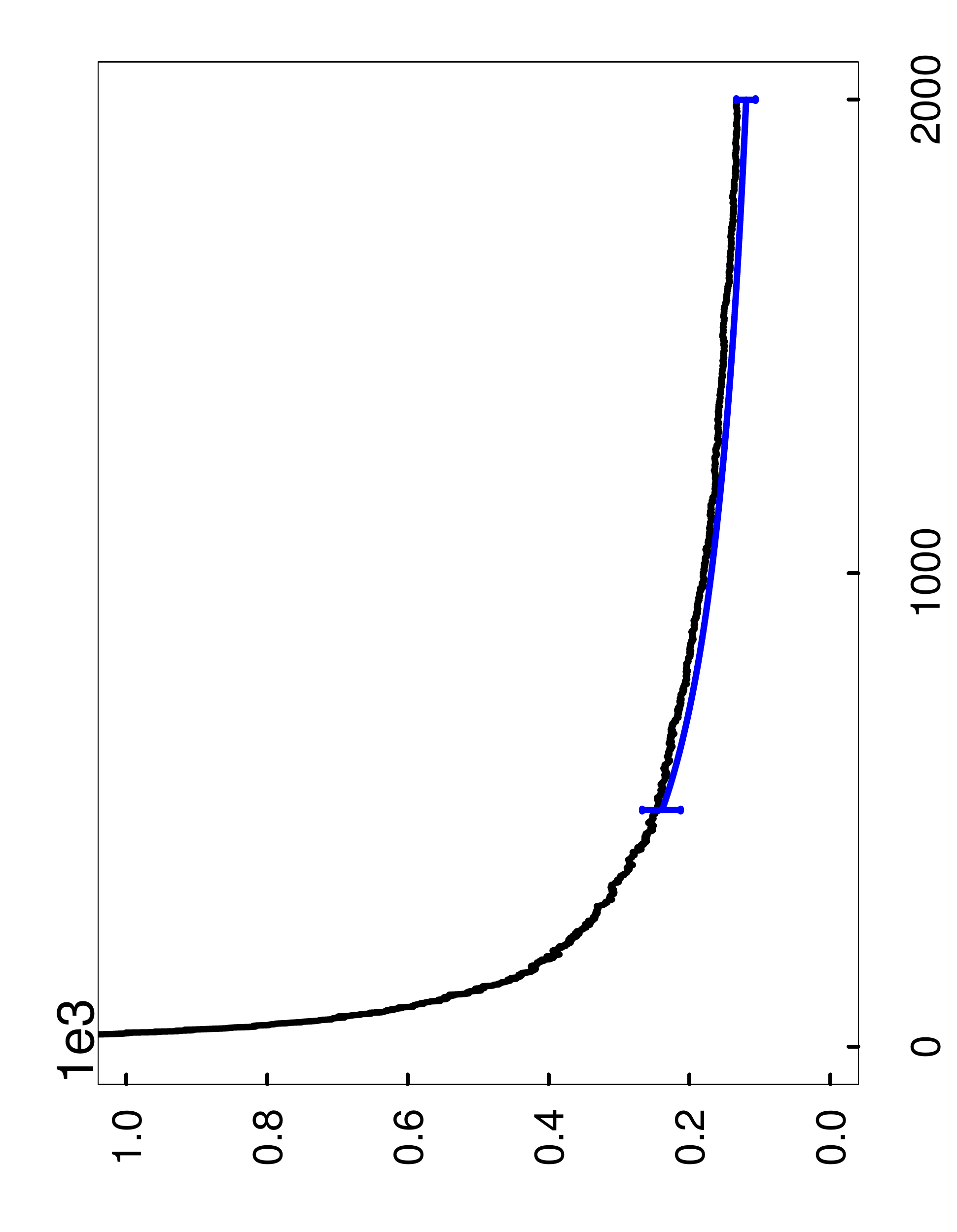} 
			\put(36,-5){\color{black}{\small ensemble size $t$}}   
			\put(-8,45){\rotatebox{90}{\small $\tq_{0.9}(t)$}}
			\put(40,65){\color{black}{\Huge -}}
			\put(48,68){\color{black}{\small $\tq_{0.9}(t)$ (true)}}
			\put(40,58){\color{blue}{\Huge -}}
			\put(48,61){\color{black}{\small estimated}}
		\end{overpic}
				\vspace{0.2cm}
		\caption{Housing Data}\label{fig:housingvi}
	\end{minipage}
	\begin{minipage}[t]{0.48\textwidth}
		\centering
		\begin{overpic}[width=50mm,height=50mm,angle=270]{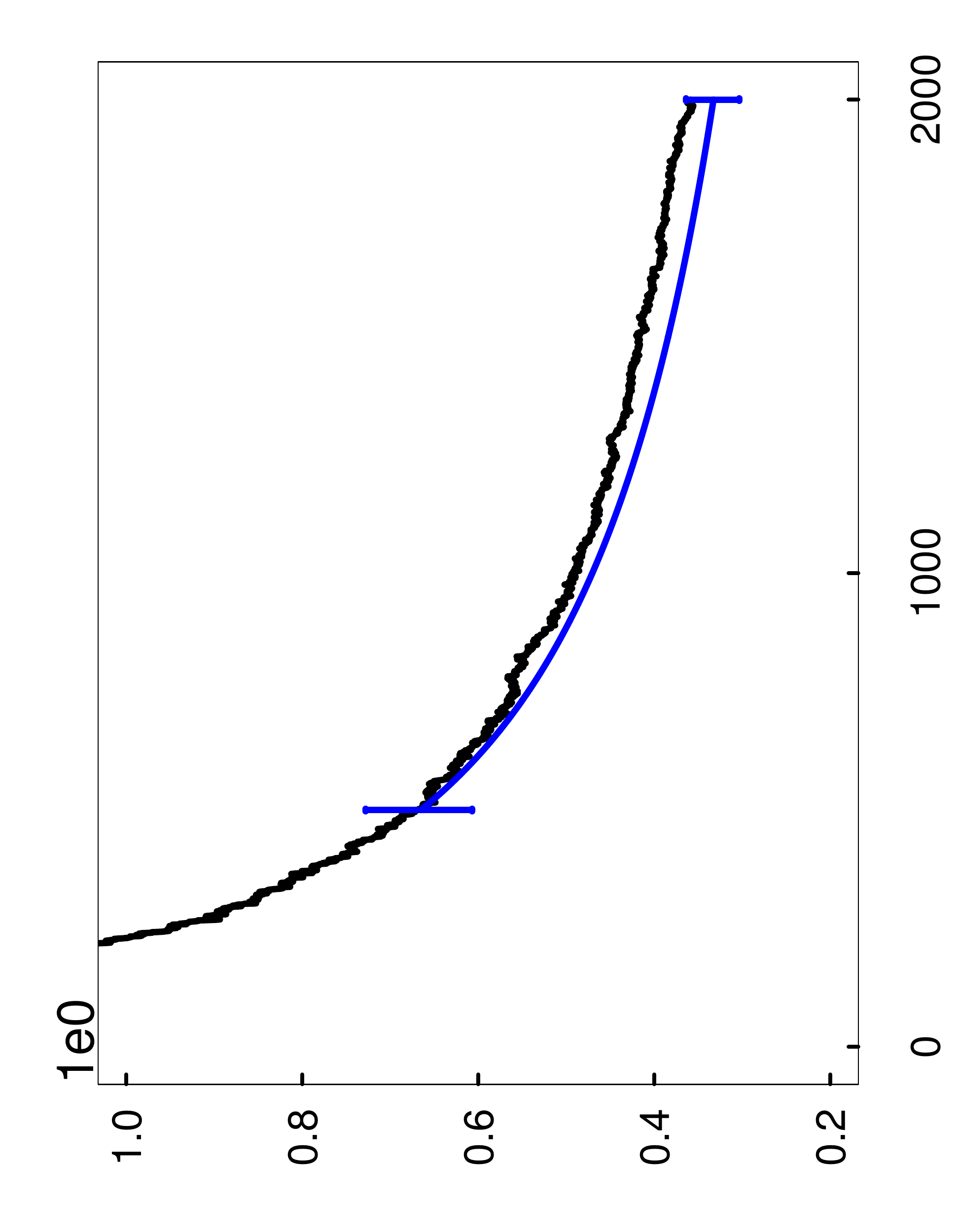} 
			\put(36,-5){\color{black}{\small ensemble size $t$}}   
			\put(-8,45){\rotatebox{90}{\small $\tq_{0.9}(t)$}}
			\put(40,65){\color{black}{\Huge -}}
			\put(48,68){\color{black}{\small $\tq_{0.9}(t)$ (true)}}
			\put(40,58){\color{blue}{\Huge -}}
			\put(48,61){\color{black}{\small estimated}}
		\end{overpic}
				\vspace{0.2cm}
		\caption{Protein Data}\label{fig:proteinvi}
	\end{minipage}
\end{figure}

\begin{figure}[htbp]
	\centering
	\begin{minipage}[t]{0.48\textwidth}
		\centering
		\begin{overpic}[width=50mm,height=50mm,angle=270]{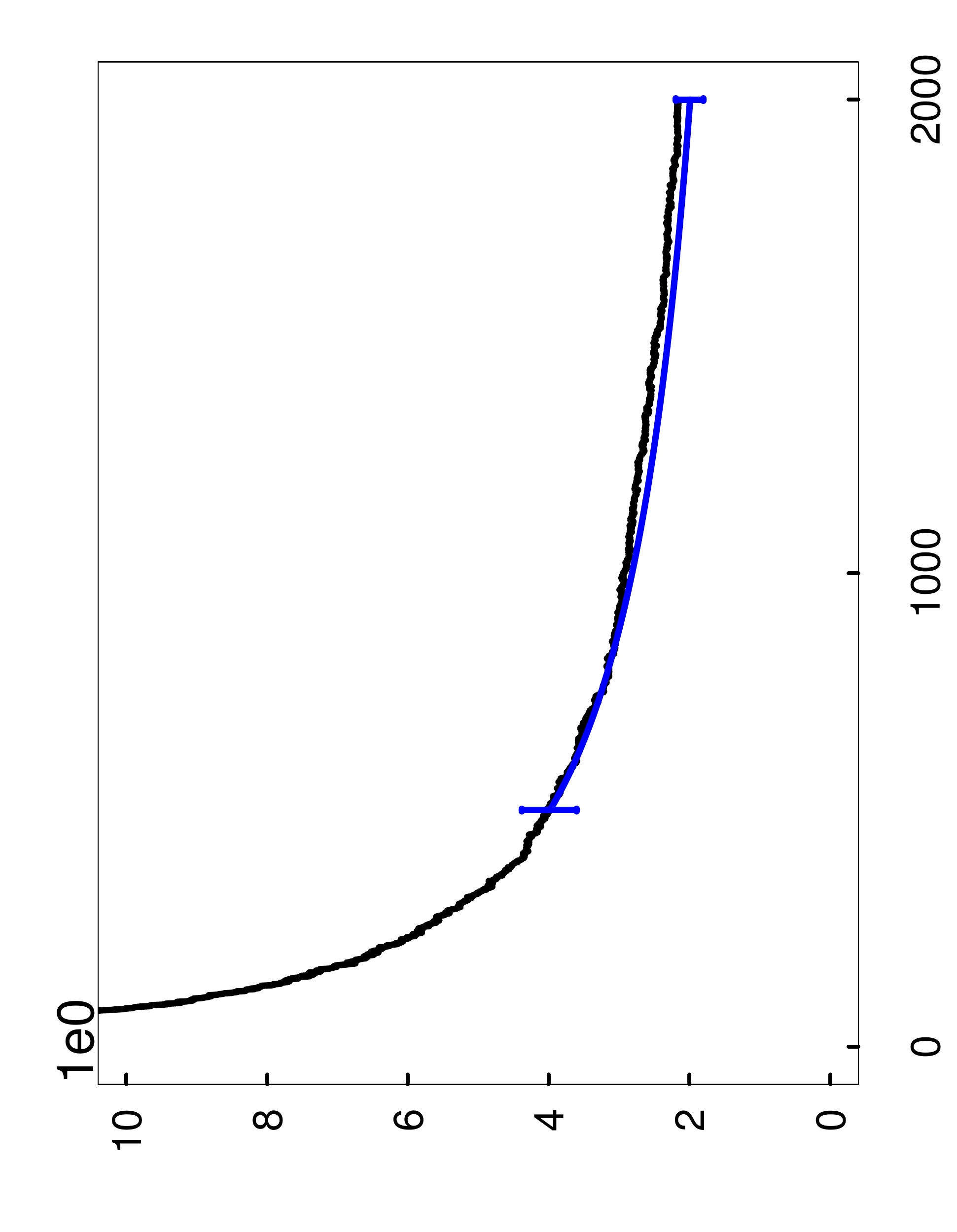} 
			\put(36,-5){\color{black}{\small ensemble size $t$}}   
			\put(-8,45){\rotatebox{90}{\small  $\tq_{0.9}(t)$}}
			\put(40,65){\color{black}{\Huge -}}
			\put(48,68){\color{black}{\small $\tq_{0.9}(t)$ (true)}}
			\put(40,58){\color{blue}{\Huge -}}
			\put(48,61){\color{black}{\small estimated}}
		\end{overpic}
				\vspace{0.2cm}
		\caption{Music Data}\label{fig:musicvi}
	\end{minipage}
	\begin{minipage}[t]{0.48\textwidth}
		\centering
		\begin{overpic}[width=50mm,height=50mm,angle=270]{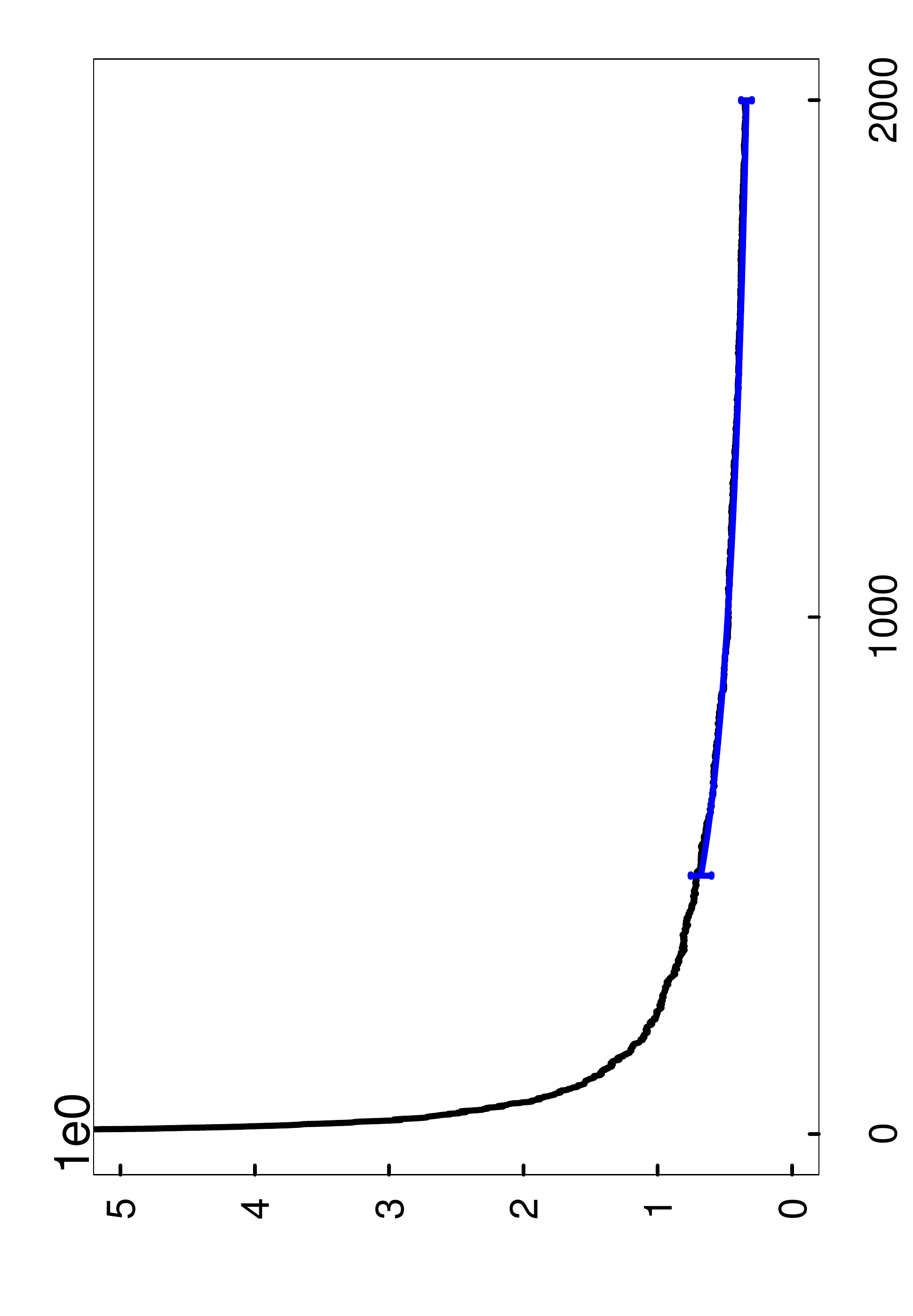} 
			\put(36,-5){\color{black}{\small ensemble size $t$}}   
			\put(-8,45){\rotatebox{90}{\small $\tq_{0.9}(t)$}}
			\put(40,65){\color{black}{\Huge -}}
			\put(48,68){\color{black}{\small $\tq_{0.9}(t)$ (true)}}
			\put(40,58){\color{blue}{\Huge -}}
			\put(48,61){\color{black}{\small estimated}}
		\end{overpic}
				\vspace{0.2cm}
		\caption{Diamond Data}\label{fig:diamondvi}
	\end{minipage}
\end{figure}

The results in the setting of variable importance are simpler to describe, since there is only one type of estimate for $\tq_{.90}(t)$. In Figures~\ref{fig:housingvi} through~\ref{fig:diamondvi}, we plot the average of the 1,000 realizations of the estimates using a blue curve, while the error bars at the endpoints represent the 10\% and 90\% empirical quantiles of the estimates. In addition, the extrapolation procedure was based on an initial ensemble size of $t_0=500$, as in the previous subsection.
From the four plots, it is clear that the extrapolated estimate displays excellent overall performance, with its bias and variance both being very small.


\bibliography{rf_reg_bib}

\newpage

\begin{center}
\Large{Supplementary Material:}\\[0.2cm]
\normalsize{
Measuring the Algorithmic Convergence of Randomized Ensembles:\\The Regression Setting}\\[0.4cm]

Miles E. Lopes \ \ \  Suofei Wu \ \ \  Thomas C. M. Lee
\end{center}


\appendix

\paragraph{Outline of proofs.} The key points of the proof of Theorem~\ref{thm:main} are explained in Appendix~\ref{app:first}, and the primary lemmas are given in Appendix~\ref{app:second}. These lemmas rely on secondary technical results and background facts which are given in Appendices~\ref{app:third} and~\ref{app:fourth} respectively.

\paragraph{Notation and conventions.} To simplify presentation, letters such as $c, c_0, c_1,$ etc., will be re-used to refer to positive absolute constants, not depending on $t$, $B$, or $k$, and likewise, these letters may take a different value at each occurrence. Regarding the quantity $\delta_{t,k,B}(\D)$ defined in equation~\eqref{eqn:deltadef} of Theorem~\ref{thm:main}, we will omit the subscripts and write $\delta(\D)$ in order to lighten notation. In addition, if $C\geq 1$ is an absolute constant, we may assume without loss of generality that
\begin{equation}\label{eqn:deltasmall}\tag{A0.1}
\delta(\D)< \ts\frac{1}{C},
\end{equation}
 because if the constant $c_0$ in Theorem~\ref{thm:main} is chosen to satisfy $c_0\geq C$, then the result is clearly true when $\delta(\D)\geq \frac{1}{C}$. Next, we will often make use of the following basic moment relations involving quantities defined on page~\pageref{momentpage},
\begin{equation}\label{eqn:moments}\tag{A0.2}
\begin{split}
\sigma(\D) & \ \leq \ 2\beta_1(\D),\\[0.2cm]
\beta_{\ell}(\D) & \ \leq \ \beta_m(\D) \text{ \ \ \ \ whenever \ \ \ \ } 1\leq \ell \leq m,\\[0.2cm]
\big(\E[|\zeta|^m|\D]\big)^{1/m} &  \ \leq \, \, 2\beta_m(\D) \text{ \ \ \ \ for any \ \ \  \ \ \ } m\geq 1.
\end{split}
\end{equation}
These relations are straightforward to verify using the Cauchy-Schwarz and Jensen inequalities, and hence the details are omitted. Furthermore, under the above condition $\delta(\D)<\frac{1}{C}$, these relations and the definition of $\delta(\D)$ in~\eqref{eqn:deltadef} imply that 
\begin{equation}\label{eqn:kt}\tag{A0.3}
k^2\leq \ts\sqrt{t},
\end{equation}
 which will be useful in simplifying some expressions. Next, recall that the quantile function $G^{-1}$ associated with a generic distribution function $G$ is defined  as 
 $$G^{-1}(r)=\inf\big\{s\in\R \, \big| \, G(s)\geq r\big\},$$
 for any $r\in(0,1)$. Lastly, the supremum norm of a function $h:\R\to\R$ is written as $\|h\|_{\infty}=\sup_{s\in\R}|h(s)|$.

\section{High-level proof of Theorem~\ref{thm:main}}\label{app:first}
Define the following distribution functions at any $s\in\R$,
\begin{align}
 F(s) &:=\P\Big(\sqrt t(\Mse_t-\mse_{\infty}) \leq s \Big| \D\Big)
\end{align}
and
\begin{align}
\hat F(s) &:= \frac 1 B \sum_{l=1}^B 1\big\{ \sqrt t( \Mse_{t,l}^*-\Mse_t) \leq s\big\},
\end{align}
where each $\Mse_{t,l}^*-\Mse_{t}$ is an independent copy of the bootstrap sample~\eqref{eqn:bootdef0}, conditionally on $\D$ and $\bxi_t$.

In Proposition~\ref{prop:firstmain} below, we will show there is an absolute constant $c_1>0$ such that
\begin{equation}\label{eqn:crucial}
\P\bigg(\|\hat F- F\|_{\infty} > c_1\delta(\D)\,\Big|\, \D  \bigg) \leq 4e^{-k/2}+\ts\frac{2}{B^2},
\end{equation}
which is the most substantial part of the proof.
Next, recall that $q_{1-\alpha}(t)$ and $\hat q_{1-\alpha}(t)$ are defined to satisfy
\begin{equation}
\begin{split}
q_{1-\alpha}(t) &=\ts\frac{1}{\sqrt{t}}F^{-1}(1-\alpha)\\[0.2cm]
%
 \hat q_{1-\alpha}(t)&=\ts\frac{1}{\sqrt{t}}\hat F^{-1}(1-\alpha),
 \end{split}
 \end{equation}
and let $\mathcal{E}$ be an event defined by
$$\mathcal{E}=\Big\{\hat q_{1-\alpha}(t)\geq \ts\frac{1}{\sqrt{t}}F^{-1}(1-\alpha-c_1\delta(\D))\Big\}.$$
By intersecting the event $\{\Mse_t-\mse_{\infty}> \hat q_{1-\alpha}(t)\}$ with $\mathcal{E}$ and $\mathcal{E}^c$, it follows that
\begin{equation}\label{eqn:combine}
\footnotesize
\begin{split}
\P\bigg(\Mse_t-\mse_{\infty}>\hat q_{1-\alpha}(t)\bigg| \D\bigg)
&\leq \P\bigg(\Mse_t-\mse_{\infty} > \ts\frac{1}{\sqrt{t}}F^{-1}\big(1-\alpha-c_1\delta(\D)\big)\bigg|\D\bigg)+\P(\mathcal{E}^c|\D)\\[0.3cm]
&\leq \alpha+c_1\delta(\D)+\P(\mathcal{E}^c|\D).
\end{split}
\end{equation}
In turn, observe that if the event $\{\|\hat F-F\|_{\infty}\leq c_1\delta(\D)\}$ holds, then 
\begin{equation}
\begin{split}
 F\big(\sqrt t\,\hat q_{1-\alpha}(t)\big) &\geq \hat F\big(\sqrt t\,\hat q_{1-\alpha}(t)\big) - \|\hat F-F\|_{\infty}\\[0.3cm]
&\geq 1-\alpha-c_1\delta(\D),
\end{split}
\end{equation}
which implies that the event $\mathcal{E}$ contains  $\{\|\hat F-F\|_{\infty}\leq c_1\delta(\D)\}$. In other words, the bound~\eqref{eqn:crucial} implies
\begin{equation}
\P(\mathcal{E}^c|\D)\leq 4e^{-k/2}+\ts\frac{2}{B^2}.
\end{equation}
Combining this with~\eqref{eqn:combine} gives
\begin{equation}
\P\bigg(\Mse_t-\mse_{\infty}\leq\hat q_{1-\alpha}(t)\bigg| \D\bigg) \geq 1-\alpha-\Big(c_1\delta(\D)+4e^{-k/2}+\ts\frac{2}{B^2}\Big).
\end{equation}
Finally, it is clear that there is an absolute constant $c_2>0$ such that $4e^{-k/2}+2/B^2\leq c_2\delta(\D)$, and so the proof is complete.
\qed

\begin{proposition}\label{prop:firstmain}
Suppose the conditions of Theorem~\ref{thm:main} hold. Then, there is an absolute constant $c_1>0$ such that
\begin{equation}
\P\bigg(\|\hat F- F\|_{\infty} > c_1\delta(\D)\Big| \D  \bigg) \leq 4e^{-k/2}+\ts\frac{2}{B^2}.
\end{equation}
\end{proposition}

\proof 
For any fixed $s\in\R$, define the distribution function
\begin{equation}\label{eqn:Ftilde}
\tilde F(s):=\P\Big(\sqrt t(\Mse_t^*-\Mse_t)\leq s \Big | \D,\bxi_t\Big).
\end{equation}
Clearly,
\begin{equation}
\|\hat F-F\|_{\infty} \leq \|\hat F-\tilde F\|_{\infty}+\|\tilde F -F\|_{\infty}.
\end{equation}
The proof amounts to bounding the two terms on the right. To consider the first term $\|\hat F-\tilde F\|_{\infty}$, note that $\hat F$ is  the empirical distribution function based on $B$ i.i.d.~samples from $\tilde F$. Therefore, we may apply the Dvoretzky-Kiefer-Wolfowitz inequality (Lemma~\ref{lem:DKW}) conditionally on $\D$ and $\bxi_t$, and then take the expectation over $\bxi_t$ to obtain
\begin{equation}\label{eqn:easypart}
\P\Big(\|\hat F-\tilde F\|_{\infty} > \ts \sqrt{\frac{\log(B)}{B}}\, \Big| \D\Big) \ \leq \frac{2}{B^2}.
\end{equation}
Handling the second term $\|\tilde F-F\|_{\infty}$ is much more involved. To do this, we consider two random variables $Z$ and $Z^*$, to be defined later, which allow the distance $\|\tilde F-F\|_{\infty}$ to be bounded in three parts:
\begin{equation}
\|\tilde F-F\|_{\infty} \ \leq \ \|\tilde F-F_{Z^*}\|_{\infty} + \|F_{Z^*}-F_Z\|_{\infty}+\|F_Z-F\|_{\infty}.
\end{equation} 
Specifically, each of the terms on the right side will be handled in Lemmas \ref{lem:thirdclt}, ~\ref{lem:compare}, and~\ref{lem:firstclt} respectively. Combining the results of those lemmas shows that there is an absolute constant $c>0$ such that
\begin{equation}\label{eqn:hardpart}
\P\bigg(\|\tilde F- F\|_{\infty} > \ts\frac{ck^2}{\sqrt{t}}\Big(\ts\frac{\beta_{3k}(\D)}{\sigma(\D)}\Big)^3 +ce^{-k/2}\,\bigg|\, \D\bigg) \ \leq 4e^{-k/2}.
\end{equation}
Finally, the proof is completed by combining the inequalities~\eqref{eqn:easypart} and~\eqref{eqn:hardpart}.\qed

\section{Primary lemmas }\label{app:second}

This section contains the three essential lemmas for proving Proposition~\ref{prop:firstmain}.
\begin{lemma}\label{lem:firstclt}
Suppose that the conditions of Theorem~\ref{thm:main} hold. Let $Z$ be a Gaussian random variable generated conditionally on $\D$ as $Z\sim N(0,\sigma^2(\D))$. Also, for any $s\in\R$, define $F_Z(s)=\P(Z\leq s\, |\, \D)$. Then, there is an absolute constant $c>0$, such that 
	\begin{equation}\label{eqn:mseclt}
	\big\| F-F_Z \big\|_{\infty} \ \leq \ \ts\frac{ck^2}{\sqrt t}\Big(\ts\frac{ \beta_{3k}(\D)}{\sigma(\D)}\Big)^3+e^{-k}.
	\end{equation}
\end{lemma}
\proof  A bit of algebra gives the relation
\begin{equation*}
\begin{split}
\sqrt{t}(\Mse_t-\mse_{\infty})= Z_t+R_t
\end{split}
\end{equation*}
where we define the random variables
\begin{align}
Z_t&:=2\sqrt{t}\,\langle \vartheta-y,\bar{T}_t-\vartheta\rangle\label{eqn:Ztdef}\\[0.3cm]
R_t &:= \sqrt{t}\,\|\bar{T}_t-\vartheta\|_{L_2}^2.\label{eqn:Rtdef}
\end{align}
Also, for each $i\in\{1,\dots,t\}$, define the random variable
\begin{equation}\label{eqn:newzetaidef}
\zeta_i := 2\,\langle \vartheta -y, T_i-\vartheta\rangle,
\end{equation}
which differs from the previous definition of $\zeta$ in~\eqref{eqn:zetaidef} only through the dependence on $T_i$.
The proof consists in showing that $Z_t$ can be approximated by a Gaussian distribution, and that $R_t$ is negligible. Observe that $Z_t$ can be written as
$$Z_t=\frac{1}{\sqrt{t}}\sum_{i=1}^t \zeta_i,$$
and note that the summands $\zeta_1,\dots,\zeta_t$ are centered, and are i.i.d.~conditionally on $\D$. If we define $F_{Z_t}(s)=\P(Z_t\leq s|\D)$ for any $s\in\R$, then Lemma~\ref{lem:approx} implies that the following inequality holds any $r>0$,
\begin{equation}\label{eqn:threebound}
\begin{split}
\big\|F-F_Z\big\|_{\infty} \ \leq \
 3\big\|F_{Z_t}-F_Z\big\|_{\infty}
\ + \ \ts\frac{2r}{\sqrt{2\pi}\sigma(\D)}
\ + \ \P(R_t\geq r|\D),
\end{split}
\end{equation}
where we note that $R_t$ is non-negative.
Hence, it remains to bound the first and third terms on the right side, and then select a value of $r$. The first term satisfies the Berry-Esseen bound
\begin{equation}
\big\|F_{Z_t}-F_Z\big\|_{\infty} \leq \ts\Big(\frac{\rho(\D)}{\sigma(\D)}\Big)^3\frac{1}{\sqrt{t}}
\end{equation}
where  $\rho(\D):=(\E[|\zeta_1|^3|\D])^{1/3}$.
Next, the third term $\P(R_t>r|\D)$ is handled in Lemma~\ref{lem:Rt}, which shows that if we take
$$r=\ts\frac{c k^2 \beta_{k}(\D)}{\sqrt t},$$
for some absolute constant $c>0$, then
$$\P(R_t\geq r| \, \D) \ \leq e^{-k}. $$
Combining the three previous bounds gives
	\begin{equation}\label{eqn:complicated}
	\big\| F-F_Z \big\|_{\infty}\ \leq \ \ts\frac{c}{\sqrt t}\bigg(\Big(\frac{\rho(\D)}{\sigma(\D)}\Big)^3+ \ts\frac{ k^2 \beta_{k}(\D)}{\sigma(\D)}\bigg)+e^{-k}.
	\end{equation}
Finally, we use the following bounds from~\eqref{eqn:moments}
	\begin{equation}
\rho(\D) \leq 2\beta_{3k}(\D) \ \ \ \ \text{ and } \ \ \  \ \beta_k(\D)\leq \beta_{3k}(\D),
	\end{equation}
 and then the stated result follows from~\eqref{eqn:complicated} after simplifying.\qed

\paragraph{Remark.} For the statement and proof of the next lemma, define the random variables
$$\zeta_i^* \ := \ 2\,\langle\bar T_t-y,T_i^*-\bar T_t\rangle,$$
for each $i\in\{1,\dots,t\}$, which are conditionally i.i.d.~given $(\D,\bxi_t)$, with mean zero. Likewise, define the moments
\begin{align}
\hat \sigma(\D,\bxi_t)^2 & \ := \ \E[(\zeta_1^*)^2|\D,\bxi_t] \ = \ \ts\frac{1}{t}\tsum_{i=1}^t\big(2\langle \bar T_t-y,T_i-\bar T_t\rangle\big)^2\label{eqn:hatsigmadef}\\[0.2cm]
\hat\rho(\D,\bxi_t)^3  & \ := \ \E[|\zeta_1^*|^3|\D,\bxi_t] \ = \ \ts\frac{1}{t}\tsum_{i=1}^t\big|2\langle \bar T_t-y,T_i-\bar T_t\rangle\big|^3.\label{eqn:hatrhodef}
\end{align}
Lastly, recall that $\tilde F$ is the distribution function of $\sqrt{t}(\Mse_t^*-\Mse_t)$ given $(\D,\bxi_t)$, as defined in~\eqref{eqn:Ftilde}.

\begin{lemma}\label{lem:thirdclt}
	Suppose that the conditions of Theorem~\ref{thm:main} hold. Let $Z^*$ be a Gaussian random variable, generated conditionally on $\D$ and $\bxi_t$ according to $Z^*\sim N(0,\hat\sigma^2(\D,\bxi_t))$. Also, for any $s\in\R$, let $F_{Z^*}(s) =\P(Z^*\leq s|\D,\bxi_t)$. Then, there is an absolute constant $c>0$, such that
	\begin{equation}\label{eqn:msecltboot}
	\P\bigg(\big\| \tilde F-F_{Z^*} \big\|_{\infty}\geq \ts\frac{ck^2}{\sqrt t}\Big(\ts\frac{\beta_{3k}(\D)}{\sigma(\D)}\Big)^3+ce^{-k/2} \bigg|\D\bigg) \ \leq 3e^{-k/2}.
	\end{equation}
\end{lemma}
\proof  The proof can be viewed as the bootstrap counterpart to the proof of Lemma~\ref{lem:firstclt}. It is straightforward to verify the relation
\begin{equation*}
\begin{split}
\sqrt{t}(\Mse_t^*-\Mse_t)= Z_t^*+R_t^*,
\end{split}
\end{equation*}
where we define the random variables
\begin{align}
Z_t^*&:=2\sqrt{t}\,\langle \bar T_t-y,\bar{T}_t^*-\bar T_t\rangle,
\label{eqn:Ztstardef}\\[0.3cm]
R_t^* &:= \sqrt{t}\,\|\bar{T}_t^*-\bar T_t\|_{L_2}^2.\label{eqn:Rtstardef}
\end{align}
Also, observe that $Z_t^*$ can be written as
$$Z_t^*=\frac{1}{\sqrt{t}}\sum_{i=1}^t \zeta_i^*.$$
Next, for any $s\in\R$, define the conditional distribution function
\begin{equation}
\begin{split}
F_{Z_t^*}(s) = \P\big(Z_t^*\leq s|\D,\bxi_t).
\end{split}
\end{equation}
In turn, Lemma~\ref{lem:approx} gives  the following bound for any realization of $\D$ and $\bxi_t$, and any fixed $r>0$,
\begin{equation}\label{eqn:threeparttemp}
\|\tilde F-F_{Z^*}\|_{\infty} \ \leq \ 3\|F_{Z_t^*}-F_{Z^*}\|_{\infty}+\ts\frac{2r}{\sqrt{2\pi}\hat\sigma(\D,\bxi_t)}+\P\big(R_t^*\geq r\big |\D,\bxi_t\big).
\end{equation}
 The first term on the right satisfies the Berry-Esseen bound,
\begin{equation}
\big\|F_{Z_t^*}-F_{Z^*}\big\|_{\infty} \leq \Big(\ts\frac{\hat \rho(\D,\bxi_t)}{\hat \sigma(\D,\bxi_t)}\Big)^3\frac{1}{\sqrt{t}}.
\end{equation}
Furthermore, the quantities $\hat \rho(\D,\bxi_t)$ and $\hat \sigma(\D,\bxi_t)$ can be controlled with the help of the following tail bounds, which are direct consequences of Lemmas~\ref{lem:rhobound} and~\ref{lem:sighatconc},
$$\P\bigg(\hat \rho(\D,\bxi_t)\leq c\beta_{3k}(\D) \, \bigg|\, \D\bigg) \ \geq \ 1- e^{-k},$$
and
$$\P\bigg(\hat \sigma(\D,\bxi_t) \geq c_1\sigma(\D)\, \bigg| \, \D\bigg) \ \geq \ 1- e^{-k}.$$
Next, to use an alternative notation for the third term on the right side of~\eqref{eqn:threeparttemp}, let 
$$\pi_r(\D,\bxi_t):=\P(R_t^*\geq r|\D,\bxi_t),$$
and also write its expectation with respect to $\bxi_t$ as
$$\pi_r(\D):=\E[\pi_r(\D,\bxi_t)|\D] =\P(R_t^*\geq r|\D).$$ 
Then, Markov's inequality gives
$$\P\bigg(\pi_r(\D,\bxi_t) \ \geq \sqrt{\pi_r(\D)}\, \bigg|\, \D\bigg) \leq \sqrt{\pi_r(\D)}.$$
In Lemma~\ref{lem:bootRt}, we show that if $r$ is chosen as
$$r=\ts\frac{c k^2 \beta_{2k}(\D)}{\sqrt t}$$
for a sufficiently large absolute constant $c>0$, then the bound
$$\pi_r(\D)\leq e^{-k}$$
holds for any realization of $\D$. Combining the ingredients above, we have
$$ \P\bigg(\big\| \tilde F-F_{Z^*} \big\|_{\infty}\geq \ts\frac{c}{\sqrt t}\Big(\Big(\ts\frac{\beta_{3k}(\D)}{\sigma(\D)}\Big)^3+\ts\frac{k^2 \beta_{2k}(\D)}{\sigma(\D)}\Big)+ce^{-k/2} \bigg|\D\bigg) \ \leq 3e^{-k/2}.$$
Finally, the term involving $1/\sqrt{t}$ can be simplified by making use of the simple inequalities 
$$\sigma(\D)\leq 2\beta_{2k}(\D)\leq 2\beta_{3k}(\D).$$ This leads to the stated result.\qed

\begin{lemma}\label{lem:compare}
Suppose that the conditions of Theorem~\ref{thm:main} hold. Let $F_Z$ and $F_{Z^*}$ be as defined in the statements of Lemmas~\ref{lem:firstclt} and~\ref{lem:thirdclt}. Then, there is an absolute constant $c>0$ such that 
\begin{equation}
\P\bigg( \|F_Z-F_{Z^*}\|_{\infty}  \geq  \ts\frac{ck}{\sqrt t}\Big(\frac{\beta_{2k}(\D)}{\sigma(\D)}\Big)^{\!2}\, \bigg| \, \D\bigg)  \ \leq e^{-k}.
\end{equation}
\end{lemma}

\proof
Recall that $F_Z$ and $F_{Z^*}$ correspond to centered Gaussian distributions. It is a basic fact about the function $\Phi$ that the following bound holds for any positive numbers $\sigma_1$ and $\sigma_2,$
\begin{equation}
\sup_{s\in\R}\Big|\Phi(\ts\frac{s}{\sigma_1})-\Phi(\ts\frac{s}{\sigma_2})\Big| \ \leq \ c\Big|\ts\frac{\sigma_2^2}{\sigma_1^2}-1\Big|,
\end{equation}
where $c>0$ is an absolute constant. Since the respective variances of $F_Z$ and $F_{Z^*}$ are $\sigma^2(\D)$ and $\hat\sigma^2(\D,\bxi_t)$, this means
$$ \|F_Z-F_{Z^*}\|_{\infty} \leq c\Big|\ts\frac{\hat\sigma^2(\D,\bxi_t)}{\sigma^2(\D)}-1\Big|.$$
Combining this inequality with Lemma~\ref{lem:sighatconc} (below) completes the proof.
\qed

\section{Secondary lemmas}\label{app:third}

\paragraph{Remark.} Recall that $R_t$ is defined in~\eqref{eqn:Rtdef} as $R_t=\sqrt t \|\bar T_t-\vartheta\|_{L_2}^2.$

\begin{lemma}\label{lem:Rt}
Suppose the conditions of Theorem~\ref{thm:main} hold. Then, there is an absolute constant $c>0$, such that 
$$\P\bigg(R_t\geq \ts\frac{c k^2 \beta_{k}(\D)}{\sqrt t}\, \bigg| \, \D\bigg) \ \leq e^{-k}. $$
\end{lemma}
\proof 
The proof is based on the inequality
\begin{equation}\label{eqn:cheby}
\P(R_t \geq s|\D) \leq \frac{\E[R_t^k|\D]}{s^k},
\end{equation}
with a suitably chosen number $s>0$.
In order to control $\E[R_t^k|\D]$, we will use a version of Rosenthal's inequality that is applicable to sums of independent Banach-valued random variables, as given in Lemma~\ref{lem:talagrand}. Specifically, this lemma shows that
\begin{equation}\label{eqn:talagrand}
\footnotesize
\Big(\E\big[\|\bar T_t- \vartheta\|_{L_2}^{2k}\big|\D\big]\Big)^{\frac{1}{2k}} \ \leq c k\bigg\{\Big(\E\big[\|\bar T_t- \vartheta\|_{L_2}^2\big|\D\big]\Big)^{1/2}+\Big(\E\Big[\tsum_{i=1}^t \|\ts\frac{1}{t}(T_i-\vartheta)\|_{L_2}^{2k}\Big|\D\Big]\Big)^{\frac{1}{2k}}\bigg\},
\end{equation}
where $c>0$ is an absolute constant. Regarding the first term on the right, we may use the fact that $T_1,\dots,T_t$ are conditionally i.i.d.~given $\D$ to obtain
\begin{equation}
\begin{split}
 \Big(\E\big[\|\bar T_t- \vartheta\|_{L_2}^2\big|\D\big]\Big)^{1/2} & \, = \, \ts\frac{1}{\sqrt t}\Big(\E\big[\|T_1- \vartheta\|_{L_2}^2\big|\D\big]\Big)^{1/2} \\[0.3cm]
&\, \leq \, \ts\frac{\sqrt{\beta_1(\D)}}{\sqrt t}.
\end{split}
\end{equation}
The second term on the right side of~\eqref{eqn:talagrand} can be bounded as
\begin{equation}
\begin{split}
\Big(\tsum_{i=1}^t\E\big[ \|\ts\frac{1}{t}(T_i-\vartheta)\|_{L_2}^{2k}\big|\D\big] \Big)^{\frac{1}{2k}} & \leq \ts t^{-1} \cdot t^{\frac{1}{2k}}\cdot\sqrt{\beta_k(\D)}\\[0.2cm]
&\leq \ts\frac{\sqrt{\beta_{k}(\D)}}{\sqrt t}.
\end{split}
\end{equation}
 Recalling the prefactor of $\sqrt{t}$ in the definition of $R_t$, as well as the fact that $\beta_1(\D)\leq \beta_{k}(\D)$, it follows that the previous work can be combined as
\begin{equation}
\begin{split}
\big(\E\big[R_t^k\big|\D\big]\big)^{1/k} &= t^{1/2} \big(\E\big[\|\bar T_t-\vartheta\|_{L_2}^{2k}\big|\D\big]\big)^{1/k}\\[0.3cm]
& \ \leq \ \ts\frac{c k^2 \beta_{k}(\D)}{\sqrt t}.
\end{split}
\end{equation}
Hence, if we take 
$$s= e\cdot \ts\frac{c k^2 \beta_{k}(\D)}{\sqrt t}$$
in the inequality~\eqref{eqn:cheby}, then the proof is complete.\qed

\begin{lemma}\label{lem:sighatconc}
Suppose that the conditions of Theorem~\ref{thm:main} hold. Then, there are absolute constants $c_0,c_1>0$ such that
\begin{equation}\label{eqn:firstsighat}
\P\bigg(\ts\big|\frac{\hat\sigma^2(\D,\bxi_t)}{\sigma^2(\D)}-1\big| \geq \frac{c_0k}{\sqrt{t}}\big(\frac{\beta_{2k}(\D)}{\sigma(\D)}\big)^2\, \bigg|\, \D\bigg) \ \leq e^{-k},
\end{equation}
and
\begin{equation}\label{eqn:secondsighat}
\P\Big(\hat\sigma(\D,\bxi_t)\geq  c_1\sigma(\D)\Big) \ \geq 1-e^{-k}.
\end{equation}
\end{lemma}

\proof  Note that the second bound~\eqref{eqn:secondsighat} follows from the first bound~\eqref{eqn:firstsighat} due to the inequality
$$\big|\ts\frac{\hat\sigma(\D,\bxi_t)}{\sigma(\D)}-1\big|\leq\ts\big|\frac{\hat\sigma^2(\D,\bxi_t)}{\sigma^2(\D)}-1\big|,$$
as well as the condition~\eqref{eqn:deltasmall}. In order to prove~\eqref{eqn:firstsighat}, the main idea is to derive a quantity $b(\D)$ satisfying
$$\Big(\E\Big[\big|\hat\sigma^2(\D,\bxi_t)-\sigma^2(\D)\big|^k\Big|\D\Big]\Big)^{1/k} \, \leq  \, b(\D),$$
and then Chebyshev's inequality gives
\begin{equation}\label{eqn:Chebyshevtemp}
\P\bigg(\big|\hat\sigma^2(\D,\bxi_t)-\sigma^2(\D)\big| \, \geq e\cdot b(\D)\, \bigg|\, \D\bigg) \, \leq e^{-k}.
\end{equation}

To derive $b(\D)$, first recall that
$$\hat\sigma^2(\D,\bxi_t)=\E\big[(\zeta_1^*)^2\big| \D,\bxi_t\big] = \ts\frac{1}{t}\displaystyle\sum_{i=1}^t\big(2\langle \bar T_t-y,T_i-\bar T_t\rangle\big)^2.$$
Simple algebra gives the relation
 
  \begin{equation}\label{eqn:deltairelation}
2\big\langle \bar T_t-y,T_i-\bar T_t\big\rangle  = \ \zeta_i \ + \Delta_i
\end{equation}
where we put
\begin{equation}\label{eqn:Deltaidef}
\Delta_i:=2\big\langle \bar T_t-\vartheta, T_i-\vartheta\big \rangle \ + \  2\big\langle \bar T_t-y, \vartheta -\bar T_t\big\rangle.
\end{equation}
This allows $\hat\sigma^2(\D,\bxi_t)$ to be written as
$$\hat\sigma^2(\D,\bxi_t) = \frac 1t \sum_{i=1}^t \zeta_i^2 \ + \ \frac 1t \displaystyle\sum_{i=1}^t 2\zeta_i\Delta_i+\Delta_i^2,$$
and so the triangle inequality for the conditional $L_k$ norm $(\E[|\cdot|^k|\D])^{1/k}$ gives
\begin{equation}\label{eqn:sighattriangle}
\small
\Big(\E\Big[\big| \hat\sigma^2(\D,\bxi_t) -\sigma^2(\D)\big|^k \Big|\D\Big]\Big)^{1/k} \ \leq \ 
A_1(\D) +A_2(\D),
\end{equation}
where the terms on the right are defined as
\begin{align}
A_1(\D) &:= \Big(\E\Big[ \big|\ts\frac 1t \tsum_{i=1}^t (\zeta_i^2-\sigma^2(\D))\big|^k\Big|\D\Big]\Big)^{1/k}\label{eqn:A1def}\\[0.2cm]
A_2(\D) &:=  \Big(\E\Big[\big|\ts\frac 1t \tsum_{i=1}^t 2\zeta_i\Delta_i+\Delta_i^2\big|^k\Big|\D\Big]\Big)^{1/k}.\label{eqn:A2def}
\end{align}

To handle the term $A_1(\D)$, a straightforward calculation based on the bound $(\E[|\zeta_1|^k|\D])^{\frac{1}{k}}\leq 2\beta_{k}(\D)$ and Rosenthal's inequality (Lemma~\ref{lem:talagrand}) shows that 
\begin{equation}\label{eqn:var1stpiece}
A_1(\D)\ \leq \frac{c k \beta_{2k}^2(\D)}{\sqrt t},
\end{equation}
where $c>0$ is an absolute constant. Next, using the triangle and Cauchy-Schwarz inequalities, it is simple to check that the second term $A_2(\D)$ satisfies
\begin{equation}\label{eqn:A2}
A_2(\D) \leq \ 2\big(\E\big[|\zeta_1|^{2k}|\D]\big)^{\frac{1}{2k}}\big(\E\big[|\Delta_1|^{2k}\big|\D\big]\big)^{\frac{1}{2k}}+\big(\E\big[|\Delta_1|^{2k}\big|\D\big]\big)^{\frac{1}{k}}.
\end{equation}
To complete the proof, it suffices to bound the quantity $\big(\E\big[|\Delta_1|^{k}\big|\D\big]\big)^{\frac{1}{k}}$ for general $k$. 
Using steps analogous to the ones in the bound~\eqref{eqn:A2}, we obtain
\small
$$\big(\E\big[|\Delta_1|^k\big|\D\big]\big)^{\frac{1}{k}} \ \leq \ 2\big(\E\big[\|\bar T_t-\vartheta\|_{L_2}^{2k}\big|\D\big]\big)^{\frac{1}{2k}}\bigg\{\big(\E\big[\|T_1-\vartheta\|_{L_2}^{2k}\big|\D\big]\big)^{\frac{1}{2k}}+\big(\E\big[\|\bar T_t-y\|_{L_2}^{2k}\big|\D\big]\big)^{\frac{1}{2k}}\bigg\}.$$
\normalsize
Next, recall that the argument following the bound~\eqref{eqn:talagrand} in the proof of Lemma~\ref{lem:Rt} leads to
$$\Big(\E\big[\|\bar T_t-\vartheta\|_{L_2}^{2k}\big|\D\big]\Big)^{\frac{1}{2k}} \leq \ts\frac{ck\sqrt{\beta_k(\D)}}{\sqrt t}, $$
for some absolute constant $c>0$. In addition, if we apply a discrete version of Jensen's inequality  
$$\|\bar T_t-y\|_{L_2}^{2k}\ \leq \ \ts\frac{1}{t}\sum_{i=1}^t \|T_i-y\|_{L_2}^{2k},$$
and use Assumption \textbf{A2} to get
\begin{equation}
\begin{split}
\big(\E\big[\|T_i-y\|_{L_2}^{2k}\big|\D\big]\big)^{\frac{1}{2k}}&\leq \sqrt{\beta_k(\D)} \text{ \ \ \  \  \  and  }\\[0.3cm]
 \big(\E\big[\|T_1-\vartheta\|_{L_2}^{2k}\big|\D\big]\big)^{\frac{1}{2k}}&\leq \sqrt{\beta_k(\D)},
 \end{split}
 \end{equation}
 then
\begin{equation}\label{eqn:DeltaLk}
\begin{split}
\big(\E\big[|\Delta_1|^k\big|\D\big]\big)^{\frac{1}{k}}
& \ \leq \ \ts\frac{c k\beta_k(\D)}{\sqrt t}.
\end{split}
\end{equation}
To combine the work above, recall the condition~\eqref{eqn:kt}, and note that we must replace $k$ with $2k$ when relating the bound~\eqref{eqn:DeltaLk} to $\big(\E[|\Delta_1|^{2k}|\D]\big)^{\frac{1}{2k}}$. Altogether, we conclude
$$\Big(\E\Big[\big| \hat\sigma^2(\D,\bxi_t) -\sigma^2(\D)\big|^k\Big|\D\Big]\Big)^{\frac{1}{k}} \ \leq \ \ts\frac{c k\beta_{2k}^{2}(\D)}{\sqrt t}.$$
Hence, if we define $b(\D)$ to be the right side above, then the bound~\eqref{eqn:Chebyshevtemp} completes the proof.\qed

\begin{lemma}\label{lem:rhobound}
Suppose the conditions of Theorem~\ref{thm:main} hold. Then, there is an absolute constant $c>0$ such that 
$$\P\Big(\hat\rho(\D,\bxi_t) \geq c\beta_{3k}(\D) \, \Big| \,\D\Big) \leq e^{-k}.$$
\end{lemma}

\proof Recall that
\begin{equation}
\begin{split}
\hat\rho(\D,\bxi_t)^3 &=\E\big[|\zeta_1^*|^3\big| \D,\bxi_t\big]= \ts\frac{1}{t}\displaystyle\sum_{i=1}^t \big|2\langle \bar T_t-y,T_i-\bar T_t\rangle\big|^3.
\end{split}
\end{equation}
Also, if we recall the relation~\eqref{eqn:deltairelation}
  \begin{equation}
2\big\langle \bar T_t-y,T_i-\bar T_t\big\rangle  \, = \, \zeta_i + \Delta_i,
\end{equation}
with $\Delta_i$ as defined in~\eqref{eqn:Deltaidef}, then
\begin{equation}
\hat \rho(\D,\bxi_t)^3 \ \leq \ts \frac c t\displaystyle \sum_{i=1}^t (|\zeta_i|^3 +|\Delta_i|^3).\end{equation}

To derive a high probability upper bound on $\hat \rho(\D,\bxi_t)$, it is enough to use Chebyshev's inequality 
\begin{equation}\label{eqn:rhocheby}
\P\Big(\hat\rho(\D,\bxi_t)^3 \ \geq e\cdot \big(\E[\hat\rho(\D,\bxi_t)^{3k}|\D]\big)^{1/k}\Big|\D\Big)  \ \leq \ e^{-k},
\end{equation}
in conjunction with a bound on $(\E[\hat\rho(\D,\bxi_t)^{3k}|\D])^{1/k}$.  By the triangle inequality for the conditional $L_k$ norm $(\E[|\cdot|^k|\D])^{1/k}$, we have
\begin{equation}\label{eqn:rhotri}
 (\E[\hat\rho(\D,\bxi_t)^{3k}|\D])^{1/k} \ \leq \ c \big(\E[|\zeta_1|^{3k}|\D]\big)^{1/k}+c\big(\E[|\Delta_1|^{3k}|\D]\big)^{1/k}
\end{equation}
It is straightforward to check that the first term on the right satisfies
$$\big(\E[|\zeta_1|^{3k}|\D]\big)^{1/k} \ \leq \ c\beta_{3k}^3(\D).$$ 
Meanwhile, the following crude (but adequate) bound for the second term on the right side of~\eqref{eqn:rhotri} can be obtained directly from~\eqref{eqn:DeltaLk} and the condition~\eqref{eqn:kt}, 
$$\big(\E[|\Delta_1|^{3k}|\D]\big)^{1/k} \ \leq \ c\beta_{3k}^3(\D).$$
  Altogether, we have
$$ (\E[\hat\rho(\D,\bxi_t)^{3k}|\D])^{1/k} \ \leq \ c\beta_{3k}^3(\D),$$
and so the stated result follows from the Chebyshev bound~\eqref{eqn:rhocheby}. \qed

\paragraph{Remark.} Recall that $R_t^*$ is defined in equation~\eqref{eqn:Rtstardef} as
$$R_t^*=\sqrt t\|\bar T_t^*-\bar T_t\|_{L_2}^2.$$

\begin{lemma}\label{lem:bootRt}
Suppose the conditions of Theorem~\ref{thm:main} hold. Then, there is an absolute constant $c>0$ such that
$$\P\bigg(R_t^*\geq  \ts\frac{c k^2 \beta_{k}(\D)}{\sqrt t}\, \bigg| \, \D\bigg) \ \leq e^{-k}. $$
\end{lemma}

\proof 
The proof is similar to that of Lemma~\ref{lem:Rt}, and proceeds by developing a bound on the conditional moment
$$\big(\E\big[(R_t^*)^k\big|\D\big]\big)^{1/k} = t^{1/2} \, \big(\E\big[\|\bar T_t^*-\bar T_t\|_{L_2}^{2k}\big| \D\big]\big)^{1/k}.$$

To begin, note that $\bar T_t^*-\bar T_t$ is a sum of i.i.d., zero-mean~Banach-valued random variables, conditionally on $\D$ and $\bxi_t$. So, if we apply Lemma~\ref{lem:talagrand} with the inequality $(a+b)^{2k}\leq 2^{2k}(a^{2k}+b^{2k})$, then

\begin{equation}\label{eqn:talagrandstar}
\footnotesize
\E\big[\|\bar T_t^*- \bar T_t\|_{L_2}^{2k}\big|\D,\bxi_t\big] \leq  (ck)^{2k}\bigg\{\Big(\E\big[\|\bar T_t^*- \bar T_t\|_{L_2}^2\big|\D,\bxi_t\big]\Big)^{k}
\ + \ \ \tsum_{i=1}^t \E\big[\|\ts\frac{1}{t}(T_i^*-\bar T_t)\|_{L_2}^{2k}\big|\D,\bxi_t\big]\bigg\}.
\end{equation}
where $c>0$ is an absolute constant.  Direct calculation shows that the first term on the right satisfies
\begin{equation*}
\begin{split}
\Big(\E\big[\|\bar T_t^*- \bar T_t\|_{L_2}^2\big|\D,\bxi_t\big]\Big)^{k}
& =\Big(\ts\frac{1}{t}\,\E\big[\| T_1^*- \bar T_t\|_{L_2}^{2}\big|\D,\bxi_t\big]\Big)^k\\[0.2cm]
&= \ts\frac{1}{t^k}\Big(\ts\frac{1}{t}\sum_{i=1}^t\| T_i- \bar T_t\|_{L_2}^{2}\Big)^k\\[0.2cm]
&\leq \ts\frac{1}{t^{k+1}}\sum_{i=1}^t\| T_i- \bar T_t\|_{L_2}^{2k},
\end{split}
\end{equation*}
where Jensen's inequality has been used in the last step. 
Likewise, the second term in~\eqref{eqn:talagrandstar} satisfies
\begin{equation*}
\begin{split}
\tsum_{i=1}^t \E\big[\|\ts\frac{1}{t}(T_i^*-\bar T_t)\|_{L_2}^{2k}\big|\D,\bxi_t\big] 
& \ =t \, \E[\|\ts\frac{1}{t}(T_1^*-\bar T_t)\|_{L_2}^{2k}|\D,\bxi_t] \\[0.3cm]
& \ = \  \tsum_{i=1}^t \|\ts\frac{1}{t}(T_i-\bar T_t)\|_{L_2}^{2k}.
\end{split}
\end{equation*}
Hence, if we integrate with respect to $\bxi_t$, then~\eqref{eqn:talagrandstar} leads to
\begin{equation*}\label{eqn:almostdone}
 \E\big[\|\bar T_t^*- \bar T_t\|_{L_2}^{2k}\big|\D\big] \leq (ck)^{2k}\cdot \big(\ts\frac{1}{t^{k}}+\ts\frac{1}{t^{2k-1}}\big)\cdot\E\big[\|T_1-\bar T_t\|_{L_2}^{2k}\big|\D\big].
 \end{equation*}
The last factor on the right can be decomposed as
\begin{equation}
\begin{split}
\E\big[\|T_1-\bar T_t\|_{L_2}^{2k}\big|\D\big] & \ \leq \ 2^{2k}\cdot\E\big[\|T_1-\vartheta\|_{L_2}^{2k}\big|\D\big]+2^{2k}\cdot \E\big[\|\bar T_t - \vartheta\|_{L_2}^{2k}\big|\D\big]\\[0.4cm]
& \ \leq \ 2^{2k}\cdot\beta_{k}^{k}(\D)+2^{2k}\cdot\ts\frac{1}{t}\sum_{i=1}^t \E\big[\| T_i-\vartheta\|_{L_2}^{2k}\big|\D\big]\\[0.4cm]
& \ \leq \ 2^{2k+1}\cdot\beta_{k}^{k}(\D)  
\end{split}
\end{equation}
Combining the last two displays, it follows after some simplification that
\begin{equation}
\big(\E\big[(R_t^*)^k\big|\D\big]\big)^{1/k}  \leq \ts\frac{ck^2\beta_{k}(\D)}{\sqrt t},
\end{equation}
and this completes the proof by using Chebyshev's inequality in the same manner as in the proof of Lemma~\ref{lem:Rt}.\qed

\section{Background results}\label{app:fourth}

The following inequality is a modified version of the main result in~\citep{Talagrand:1989}. (See also~\citep{Zinn} and~\citep{Kwapien}.)
\begin{lemma}\label{lem:talagrand}
Let $W_1,\dots,W_m$ be independent and zero-mean elements of a Banach space with norm $\|\cdot\|$. Then, there is an absolute constant $c>0$, such that for any $r\geq 1$,
\begin{equation}
\small
\bigg(\E\Big[\big\|\tsum_{j=1}^m W_j\big\|^r\Big]\bigg)^{1/r} \ \leq \ c\, r\Bigg\{\Big(\E\Big[\big\|\tsum_{j=1}^m W_j\big\|^2\Big]\Big)^{1/2}\, + \, \Big(\sum_{j=1}^m\E\big[\|W_j\|^r\big]\Big)^{1/r}\Bigg\}.
\end{equation}
In particular, if $W_1,\dots,W_m$ are scalar random variables, and $\|\cdot\|_r=\E[|\cdot|^r])^{1/r}$, then
	\begin{equation}
	\big\|\tsum_{j=1}^m W_j\big\|_r\leq cr \bigg\{ \big\|\tsum_{j=1}^m W_j\big\|_2 + \big(\tsum_{j=1}^m \big\|W_j\|_r^r\big)^{1/r}\bigg\}.
	\end{equation}
\end{lemma}

~\\

\noindent The next lemma is the Dvoretzky-Kiefer-Wolfowitz inequality~\citep{Dvoretzky,Massart}.

\begin{lemma}\label{lem:DKW}
Let $\xi_1,\dots,\xi_m$ be independent random variables with a common distribution function $G$. Also, for any $s\in\R$, let
$$ \hat G(s)=\frac 1m \sum_{i=1}^m 1\{\xi_i\leq s\}.$$
Then, for any fixed $x>0$,
\begin{equation}
\P\bigg(\sup_{s\in\R}\big|\hat G(s)-G(s)\big| > x\bigg) \ \leq \ 2e^{-2m x^2}.
\end{equation}
\end{lemma}

\begin{lemma}\label{lem:approx}
Fix any $\tau>0$. Let $U,\, V,\, W,$ and $R$ be random variables, satisfying $U=V+R$, and $W\sim N(0,\tau^2)$. Also let $F_U,\, F_V,$ and $F_W$ denote the distribution functions of the first three variables. Then, for any $r>0$,
\begin{equation}
\begin{split}
\|F_U- F_W\|_{\infty}  \leq & \  \ 3\|F_V-F_W\|_{\infty}
\ + \ts\frac{2r}{\sqrt{2\pi} \tau}
 \ + \ \P(|R|\geq r).
\end{split}
\end{equation}
\end{lemma}

\proof It is straightforward to check that the following inequalities hold for any $s\in\R$,
$$ -\P\Big(s-|R|\leq V\leq s\Big) \ \leq \ \P(U\leq s)-\P(V\leq s) \ \leq \P\Big(s\leq V\leq s+|R|\Big) $$
and so
\begin{equation}\label{eqn:UVbound}
\Big| \P(U \leq s)-\P(V\leq s) \Big| \ \leq \ \P\Big(|V-s\, |\leq |R|\Big).
\end{equation}
Note also that for any fixed $r>0$,
$$\P\Big(|V-s|\leq |R|\,\Big) \ \leq \ \P\Big(|V-s|< r\Big)+\P\big(|R|\geq r\big).$$
Here, the first probability on the right side can be bounded as
$$\P\big(|V-s|< r\big) \ \leq \  \P\big(|W-s|\leq r\big)+ 2\|F_V-F_W\|_{\infty}.$$
Since $W\sim N(0,\tau^2)$, its distribution function is Lipschitz with parameter $\frac{1}{\sqrt{2\pi}\tau}$, and so we have
$$ \P\big(|W-s|\leq r\big) \ \leq \ \ts\frac{2r}{\sqrt{2\pi} \tau}$$
for every $s\in\R$.
Combining the last several steps with the bound~\eqref{eqn:UVbound} gives
$$\|F_U-F_V\|_{\infty} \ \leq \  2\|F_V-F_W\|_{\infty}\ + \ \ts\frac{2r}{\sqrt{2\pi}\tau} \ + \ \P(|R|\geq r).$$
In turn, adding $\|F_V-F_W\|_{\infty}$ to both sides leads to the stated bound on \mbox{$\|F_U-F_W\|_{\infty}$}.\qed
%


\end{document}